\documentclass[journal]{IEEEtran}

\usepackage{amsmath}
\usepackage{graphicx}
\usepackage{subfigure}
\usepackage{algpseudocode}
\usepackage{algorithm}
\usepackage{algorithmicx}
\usepackage{algpseudocode}
\usepackage{amssymb}
\usepackage{float}
 \usepackage{verbatim}
\usepackage{multirow}
\usepackage{makecell}
\usepackage{hyperref}
\usepackage{graphics}
\usepackage{diagbox}
\usepackage{color}
\usepackage{multirow,booktabs}
\usepackage{cleveref}

\usepackage[numbers,sort&compress]{natbib}

\begin{document}
\title{Augmenting Radio Signals with Wavelet Transform for Deep Learning-Based Modulation Recognition}

\author{Tao~Chen,
        Shilian~Zheng,
        Kunfeng~Qiu,
        Luxin~Zhang,
        Qi~Xuan,
        and Xiaoniu~Yang
\thanks{This work was supported in part by the National Natural Science Foundation of China under Grants U20B2038 and U21B2001. \textit{(Corresponding authors: Shilian Zheng; Xiaoniu Yang.)}}
\thanks{T. Chen and Q. Xuan are with the Institute of Cyberspace Security, and also with the College of Information Engineering, Zhejiang University of Technology, Hangzhou 310023, China (e-mail: ctchentao369@163.com; xuanqi@zjut.edu.cn).}
\thanks{S. Zheng, K. Qiu, L. Zhang, and X. Yang are with the No. 011 Research Center, National Key Laboratory of Electromagnetic Space Security,
Jiaxing 314033, China (e-mail: lianshizheng@126.com; yexijoe@163.com; lxzhangMr@126.com; yxn2117@126.com).}
}

\markboth{}
{Shell \MakeLowercase{\textit{et al.}}: Bare Demo of IEEEtran.cls for IEEE Journals}
\maketitle

\begin{abstract}
The use of deep learning for radio modulation recognition has become prevalent in recent years. This approach automatically extracts high-dimensional features from large datasets, facilitating the accurate classification of modulation schemes. However, in real-world scenarios, it may not be feasible to gather sufficient training data in advance. Data augmentation is a method used to increase the diversity and quantity of training dataset and to reduce data sparsity and imbalance. In this paper, we propose data augmentation methods that involve replacing detail coefficients decomposed by discrete wavelet transform for reconstructing to generate new samples and expand the training set. 
Different generation methods are used to generate replacement sequences.
Simulation results indicate that our proposed methods significantly outperform the other augmentation methods.
\end{abstract} 


\section{Introduction}

\IEEEPARstart {T}{he} increasing usage and widespread adoption of wireless communication technology have led to a more complex and diverse spectrum environment, making the allocation and management of spectrum resources a critical challenge \cite{9020304}. In this context, automatic modulation recognition technology plays a crucial role in identifying the signal types and modulation types used by various communication devices, enabling intelligent allocation and management of spectrum resources \cite{9723186,9315306}. This technology helps to ensure efficient spectrum utilization while preventing interference among different communication devices \cite{10042021,5357660,9623008}. 

The traditional modulation recognition methods mainly rely on prior knowledge to construct a likelihood function \cite{823550,7090020} or extract handcrafted features to build a classifier \cite{8891251,8685638} for recognizing the modulation types of received wireless signals. However, these traditional methods suffer from high complexity, poor robustness, and sensitivity to changes in the channel environment and interference, which leads to poor recognition ability. 
In recent years, there has been a remarkable surge in the advancement of deep learning, leading to significant breakthroughs in the domains of image analysis and natural language processing (NLP) \cite{8692836}. The main advantage of deep learning is that it can automatically learn and extract features from massive data to solve the problem of feature selection in traditional machine learning algorithms, thereby improving the performance and generalization ability of the model \cite{2014Deep}. Deep learning are also utilized in wireless communication systems to perform signal processing tasks by automatically extracting features from complex signals \cite{4599159,97080,8720067}. This application of deep learning has proven to be highly effective.


Deep learning algorithms typically require a large amount of training data to achieve good performance. However, in real-world scenarios, it may not be possible to collect sufficient training data in advance, particularly in non-cooperative communication scenarios where the receiving system needs to collect data over-the-air. This has led to a growing interest in the application of deep learning algorithms in few-shot scenarios. Data augmentation methods have been proposed to deal with this issue of limited training data. Early data augmentation methods generate new samples by flipping, rotation, clipping, and other geometric transformations on original samples \cite{9616212,9455063,9965292}. These methods can increase the diversity of data sets to a certain extent and improve the robustness and generalization performance. But these methods have some limitations, such as geometric transformation can only be carried out in a limited space. Deep learning-based data augmentation methods have evolved to use unsupervised generative adversarial networks (GANs) and their variants \cite{9701974,9531296,8319926} to address the problem of insufficient samples. In the same vein, transfer learning \cite{9084381} and meta-learning \cite{10049409} are employed to increase the training data and enhance the precision in scenarios with limited samples. Indeed, while methods such as transfer learning, meta-learning, and GANs can help mitigate the problem of insufficient data to some extent, they still require a sufficient amount of data to train effectively. In cases where the sample size is extremely small, these methods may not be enough to expand the feature space adequately and avoid overfitting of the neural network.

To address this issue, this paper investigates the use of discrete wavelet transform (DWT) for data augmentation in deep learning-based automatic modulation recognition. The wavelet transform is a time-frequency analysis technique that effectively captures the local features of signals in both time and frequency domains, and possesses the property of multi-resolution analysis. It decomposes the signal into low-frequency approximation coefficients and high-frequency detail coefficients, allowing for a more nuanced representation of the signal. The augmentation methods proposed are to replace wavelet detail coefficients for reconstruction to generate new samples and expand the training set. 
To validate the effectiveness of our proposed methods, we conduct experiments and compare them with existing augmentation methods.
Specifically, the main contributions of this paper are as follows.
\begin{itemize}
\item We propose augmentation methods that replacing wavelet detail coefficients for reconstruction to generate new samples and expand the training set. The augmentation methods we proposed include AZSR, RZSR, and RNSR. All three augmentation methods have been shown to improve performance, with RNSR outperforming AZSR and RZSR in terms of the degree of improvement. 
\item We further apply the RNSR method on IQ sequences that are decomposed and reconstructed using different wavelet base functions (RNSR-MW), which can leverage the diversity and local features of different wavelet bases to enhance the discriminative power of the generated samples and improve the performance of the modulation recognition task.
Results show that RNSR-MW method further improves performance compared to RNSR method when the number of augmentation is small.
\item We evaluate the effectiveness of RNSR-MW augmentation method across different convolutional neural network architectures and assess its performance through simulations. Simulation results demonstrate that RNSR-MW augmentation method achieves high recognition accuracy on four well-established convolutional neural network models. \item We compare RNSR-MW method with some existing augmentation methods, namely, Flip-based, SegCS3-based, SegCS3$^3$-based, and SegMC2-based methods \cite{10015697}, in order to validate its superiority. The results indicate that our proposed RNSR-MW method significantly outperforms the other methods in terms of recognition accuracy across a broad range of signal-to-noise ratios.
\end{itemize}

The rest of this paper is organized as follows. In section \uppercase\expandafter{\romannumeral2}, we introduce the related work of automatic modulation recognition and data augmentation. In section \uppercase\expandafter{\romannumeral3}, the proposed data augmentation methods are introduced in detail. Section \uppercase\expandafter{\romannumeral4} gives the simulation results. Finally, Section \uppercase\expandafter{\romannumeral5} concludes the paper.

\section{Related Work}
\subsection{Automatic Modulation Recognition}
In communication systems, the source typically emits a baseband signal with lower frequency components. However, the actual communication channel is usually band-limited. Therefore, modulation is necessary before transmitting the signal to shift the baseband signal to a high-frequency that is compatible with the channel. To perform correct demodulation and decoding at the receiving end of a communication system, it is essential to recognize the modulation types of the received signal \cite{9174643}. Traditional approaches for modulation recognition can generally be categorized into two main groups: likelihood-based (LB) methods and feature-based (FB) methods. The former relies on prior knowledge to construct a likelihood function for calculating the probability of each modulation type \cite{4698439,6176788}. Although this method is highly accurate due to its reliance on mathematical models and probability statistics theory, it requires a large amount of data for analysis, resulting in high computational complexity. 
The latter has the advantage of not requiring prior knowledge and can extract handcrafted features directly from the received signal including instantaneous features \cite{7443070,10042021}, statistical features \cite{8669002,7458322,8669647}, and time-frequency features \cite{9905063,9613195,9935275}. However, it can be sensitive to channel noise and multipath effects, which may decrease its accuracy.

In recent years, deep learning has developed rapidly and made significant progress in various fields, such as machine translation, text classification, speech recognition, and other mainstream tasks \cite{9667560,9342208,9633779}. In addition, deep learning has also been introduced into the field of communication signals including parameter estimation \cite{4599159}, direction finding \cite{97080}, and modulation recognition \cite{8720067}. Convolutional Neural Networks (CNNs) have been extensively applied in the domain of image processing, especially in tasks such as object detection and image classification. In a similar vein, CNNs have also been increasingly utilized in the field of signal processing, where they are employed to extract features in either the frequency or time domain through the use of convolution layers. For instances, the in-phase (I) and quadrature (Q) components of the samples are concatenated into a 2-dimensional data, which is then fed into a four-layer network consisting of two convolutional layers and two fully connected layers (denoted as CNN2D) for modulation recognition \cite{2016Convolutional}. In addition, state-of-the-art deep convolutional networks such as AlexNet \cite{9888752}, ResNet \cite{9625558}, and DenseNet \cite{9465520} have been successfully applied to modulation recognition tasks, demonstrating outstanding performance. Alternatively, ResNeXt \cite{9682126} has also been applied to modulation recognition tasks by introducing a new dimension of Cardinality to extract features from the time spectrum of received wireless signals. Furthermore, Recurrent Neural Networks (RNNs), such as Long Short Term Memory (LSTM) \cite{9379677} and Gate Recurrent Unit (GRU) \cite{8322633}, are commonly used in wireless communication as they can effectively handle the time-varying nature of communication signals. Similarly, graphs are also used to extract the topological structure of signals \cite{8949478,10014833}, such as the authors \cite{9695244} have introduced a novel approach called Adaptive Visibility Graph Neural Network (AvgNet) which leverages both time series characteristics of signals and the topological structure of graphs. This end-to-end modulation recognition framework automatically converts temporal signals into graphs for identifying modulation types.

\subsection{Data Augmentation}
Deep learning-based algorithms rely heavily on a large amount of training data to achieve optimal performance. However, in scenarios where communication is non-cooperative, it can be very difficult to gather sufficient number of training samples in advance. To address this issue of limited training data, data augmentation methods have been proposed. These methods involve expanding the dataset by applying transformations to the available training samples to enlarge the dataset, which can enhance the robustness of the model \cite{9943844,9885539}.


In the context of traditional signal augmentation methods, the authors in \cite{8936957} employs flipping and rotation transformations to enrich the training dataset. Additionally, they take into account the impact of the channel environment and add gaussian white noise to augment the original samples. Segment shift in cyclic (SegCS) augmentation and multiple signals concatenation (SegMC) augmentation methods proposed in \cite{10015697} expand the dataset by concatenating segments from multiple samples using different methods to improve performance. Similarly, neural networks are also utilized to generate new samples to expand the dataset. For example, an expanded dataset can be generated using generators and discriminators in a Generative Adversarial Network (GAN) \cite{9531296}, which can be utilized in all DL-based methods. The Auxiliary Classifier Generative Adversarial Networks (ACGANs) are used as generators to expand the dataset, and CNN is used as classifier to recognize the received signals. In few-shot scenarios, transfer learning \cite{9084381} and meta-learning \cite{10049409} techniques are utilized to address the challenge of inadequate training samples and achieve recognition accuracy on par with larger datasets.

\section{Method}
\subsection{Problem Formulation}
The task of modulation recognition involves recognizing the modulation type used for transmitting signals based on the received signals. When a receiver receives a baseband signal $x(n)$ through a time-varying wireless channel after modulation, the process can be described by the mathematical expression
\begin{equation}
\begin{aligned}
{r}(n)= {h}(n)* x(n) e^{j\left(2 \pi f_0n +\theta\right)}+g(n), n \in [0,L - 1],
\end{aligned}
\label{rt}
\end{equation}
where $L$ is the length of received signal $r(n)$, $*$ represents convolution operation, 
$h(n)$ is channel impulse response (CIR), $f_0$ denotes the carrier frequency offset, which arises due to factors such as Doppler shift or clock mismatch between the transmitter and receiver, $\theta$ represents the random phase deviation, $g(n)$ denotes the presence of additive white Gaussian noise (AWGN) with mean zero and variance $\sigma^2$.

Modulation recognition is essentially a pattern recognition problem. The purpose of modulation recognition is to identify the modulation type of the transmitted signal based on the received signal from the candidate modulation types, whose candidate set is
\begin{equation}
\begin{aligned}
\mathcal{M}= \{1, 2, \ldots, M\},
\end{aligned}
\label{m}
\end{equation}
where $M$ is the number of modulation types. The specific method is to maximize the probability $\rm Pr \it (M_{x(n)}\in \mathcal{M}_i \mid r(n))$ based on the received signal $r(n)$, where the notation $\rm Pr(\cdot)$ represents the calculation of probability, $M_{x(n)}$ is the modulation type of the transmitted basedband signal $x(n)$, $\mathcal{M}_i$ is the $i$-th modulation type in the candidate set. In order to facilitate subsequent processing of the received signal, the received signal is usually represented by IQ components:

\begin{equation}
\label{iq}
IQ(n)=
\left[\begin{array}{c}
I(n) \\
Q(n)
\end{array}\right]=\left[\begin{array}{c}
\operatorname{real}\left(r(n)\right) \\
\operatorname{imag}\left(r(n)\right)
\end{array}\right],
\end{equation}
where $\text{real}(\cdot)$ and $\text{imag}(\cdot)$ are used to extract the real and imaginary parts of the received signal $r(n)$, $I(n)$ and $Q(n)$ represent IQ components of the received signal $r(n)$.

Data augmentation refers to expanding a dataset by applying a series of transformations to the original samples. In the case of a signal sample $s$ with a modulation type denoted as $\ell$, the augmented dataset can be generated by applying $k$ different transformations as
\begin{equation}
\label{reiq}
\{ (s, \ell),(s_1,\ell),\ldots,(s_k,\ell)\}.
\end{equation}
Thus, by applying data augmentation methods, a single sample can be expanded into $(k+1)$ samples, effectively expanding the dataset. This augmented dataset is then utilized to train a neural network, thereby enhancing the accuracy of modulation recognition.

\subsection{Overview of the Augmentation Method}
\begin{figure*}[t]
    \centering
    \includegraphics[width=15cm]{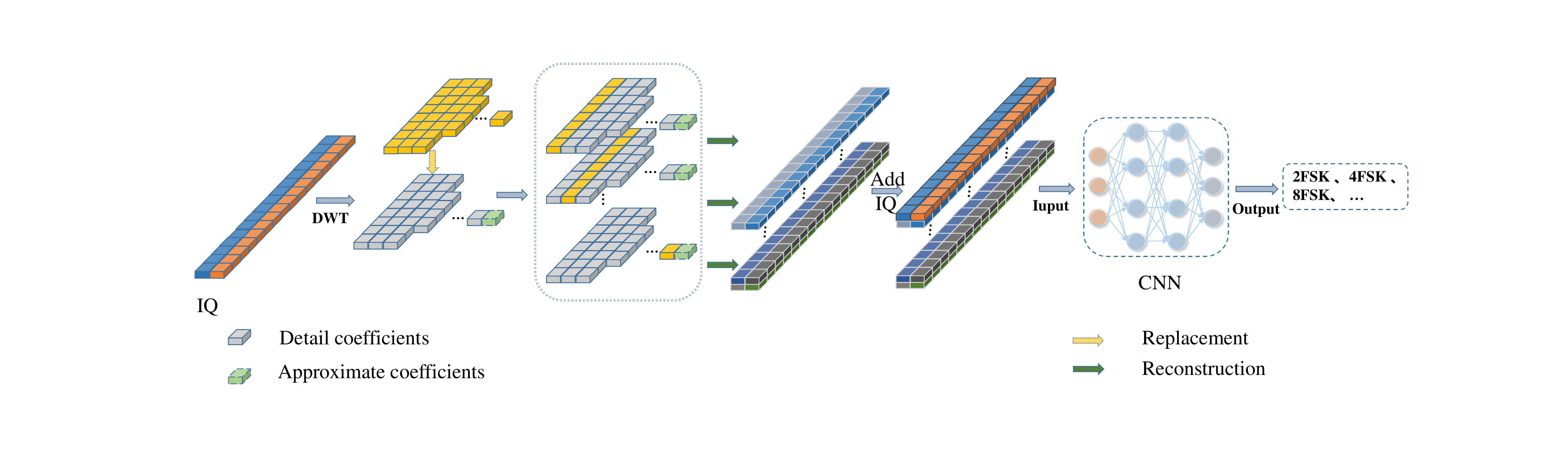}
    \caption{ The framework of data augmentation by replacing detail coefficients decomposed by discrete wavelet transform.}
    \label{Overview}
\end{figure*}
In order to address the problem of subpar performance due to the lack of sufficient samples in modulation recognition, we propose a data augmentation method for radio signals based on wavelet transform to expand the dataset and improve modulation recognition accuracy. The overall structure is shown in Fig. \ref{Overview}. Initially, we apply discrete wavelet transform to decompose the IQ sequences and obtain the approximate coefficients and detail coefficients of varying resolutions. Next, we generate a new sequence using the corresponding replacing method to replace the detail coefficients and use the replaced detail coefficients and the original approximate coefficients to reconstruct a new sample. By continuously replacing the detail coefficients through corresponding replacing methods, we can obtain a large number of new reconstructed samples. The new reconstructed samples, along with the raw IQ samples, are merged to obtain an augmented training set. Finally, this augmented training set is fed into CNNs and acquire a high-performance modulation recognition model.


In the following, we will specifically introduce the process of decomposing IQ sequences using DWT and our proposed four data augmentation methods that use different methods to generate replacement sequences for replacing detail coefficients.

\subsection{IQ Sequence Decomposition}
\subsubsection{Discrete Wavelet Transform}
Wavelet transform is a time-frequency analysis method, which can well represent the local characteristics of signals in both time and frequency domains. It has the characteristics of multi-resolution analysis.
Different from Fourier transform, wavelet transform adopts a window function whose window size is fixed but its shape can be changed, which makes wavelet transform adaptive to the signal and is very suitable for analyzing non-stationary signals \cite{6716661}. 
With the benefits mentioned above, wavelet transform has broad applications in various domains, including pattern recognition \cite{4420771}, image processing \cite{4421704}, and signal processing \cite{8407828}.

Wavelet transform includes continuous wavelet transform (CWT) and discrete wavelet transform (DWT). Because the actual sampled signal is often discrete, 
we generally use discrete wavelet transform to deal with radio signals. Wavelet transform is the wavelet coefficients of different scales obtained by a series of stretching and translation transforms on the parent wavelet $\omega(n)$, i.e. wavelet basis function. 
When wavelet transform is applied to the received signal $r(n)$, the results of wavelet transform are different due to different choice of wavelet basis functions. 
We introduce some representative wavelet basis functions as follows.

\paragraph{Haar wavelet (haar)}Haar wavelet is an orthogonal wavelet basis that is commonly utilized in image processing due to its simplicity. Nevertheless, its effectiveness is limited because of its discontinuity in the time domain. The expression of haar wavelet is 
\begin{equation}
\label{haar}
\mathcal{W}_{{haar}}(n)=\left\{\begin{array}{cc}
1 & 0 \leq n < \frac{1}{2} \\
-1 & \frac{1}{2} \leq n \leq 1 \\
0 & \text { others}
\end{array}\right..
\end{equation}

 \paragraph{Daubechies wavelet (dbN)} Daubechies wavelet has no specific mathematical expression, but its good regularity contributes to the smoothness of reconstructed signals. With higher orders (denoted as $N$), the wavelet's ability to localize frequencies improves, leading to more effective segmentation of frequency bands.

 \paragraph{Symlet wavelet (symN)} Symlet wavelet is an orthogonal wavelet that is compactly supported and approximately symmetric. It is an improved version of the db$N$ wavelet. Symlet wavelet shares similarities in continuity, support length, and filter length with the db$N$ wavelet. However, it demonstrates improved symmetry, which helps mitigate phase distortion during signal analysis and reconstruction to some extent.

 \paragraph{Coiflets wavelet (coifN)} 
Coiflet wavelet is an orthogonal wavelet function that is constructed using Daubechies wavelets. Compared to db$N$,  Coiflet wavelet has superior symmetry and includes a series of wavelets known as coif$N$ ($N$ = 1, 2, 3, 4, 5). The 2$N$ moment of coiflet wavelet function is zero, and the 2$N-1$ moment of the scale function is zero.


\paragraph{ReverseBior wavelet (rbioNr.Nd)} ReverseBior wavelet is an improvement of biorthogonal wavelet, which introduces biorthogonal wavelet to solve the incompatibility of symmetry and signal reconstruction, i,e., two dual wavelets are used respectively for signal decomposition and reconstruction. The reconstructed support range of ReverseBior wavelet is 2$Nr+1$, and the decomposed support range is 2$Nd+1$, where $Nr$ and $Nd$ are the numbers of vanishing moments of reconstruction and decomposition.

\begin{figure*}
\centering
\subfigure[]{\label{fig:subfig2:haar}
\includegraphics[width=0.17\linewidth]{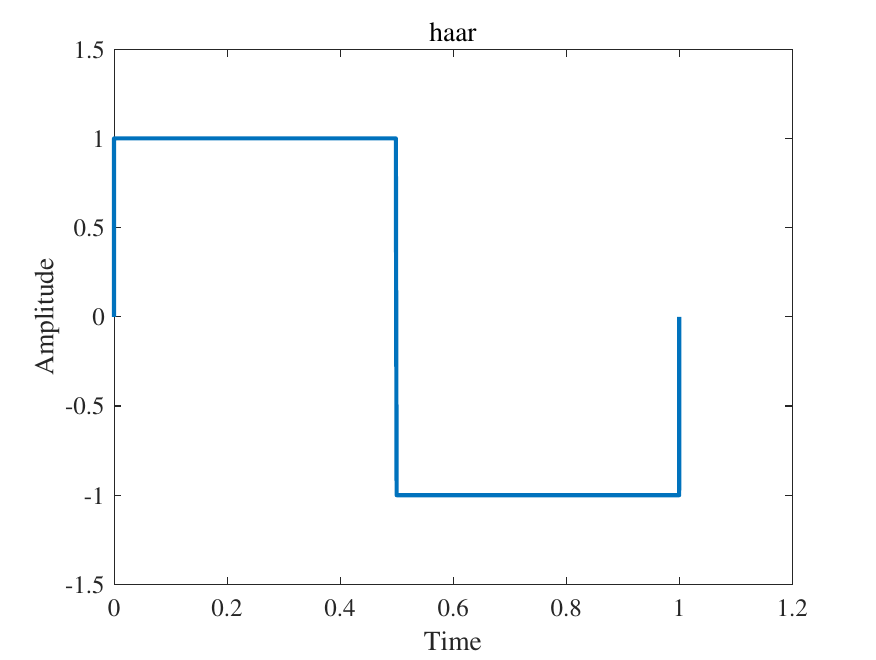}}
\subfigure[]{\label{fig:subfig2:db5}
\includegraphics[width=0.17\linewidth]{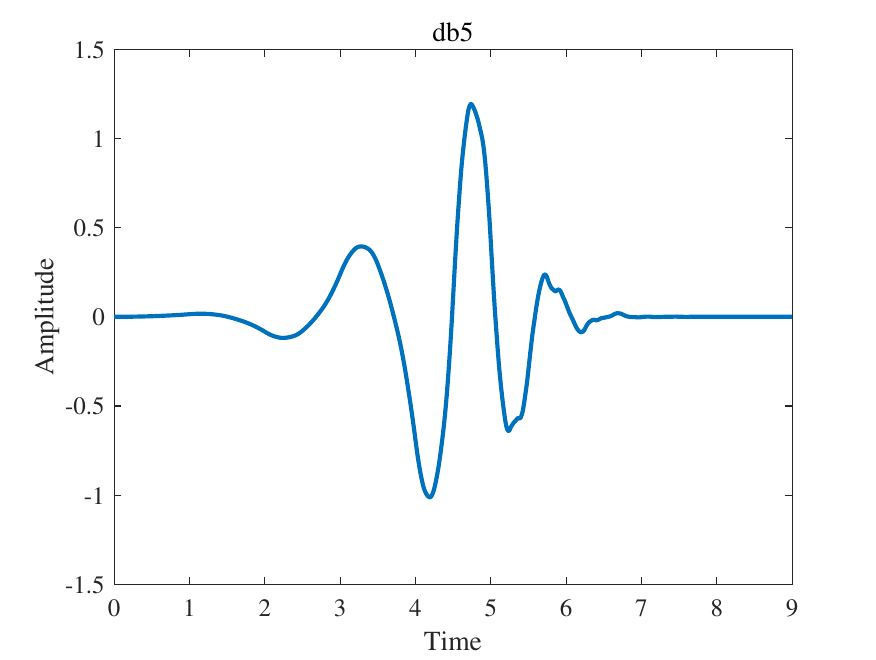}}
\subfigure[]{\label{fig:subfig2:sym5}
\includegraphics[width=0.17\linewidth]{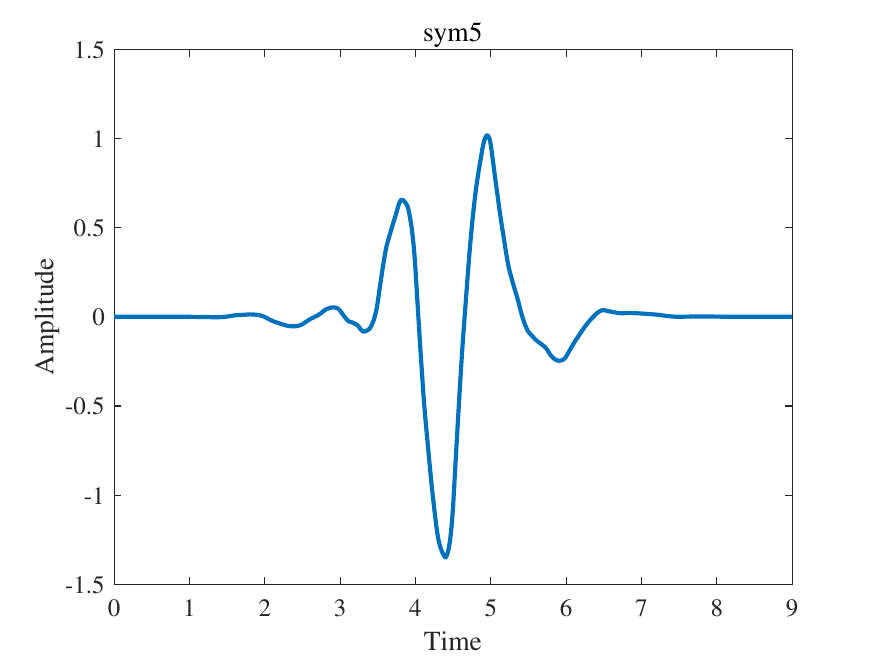}}
\subfigure[]{\label{fig:subfig2:coif3}
\includegraphics[width=0.17\linewidth]{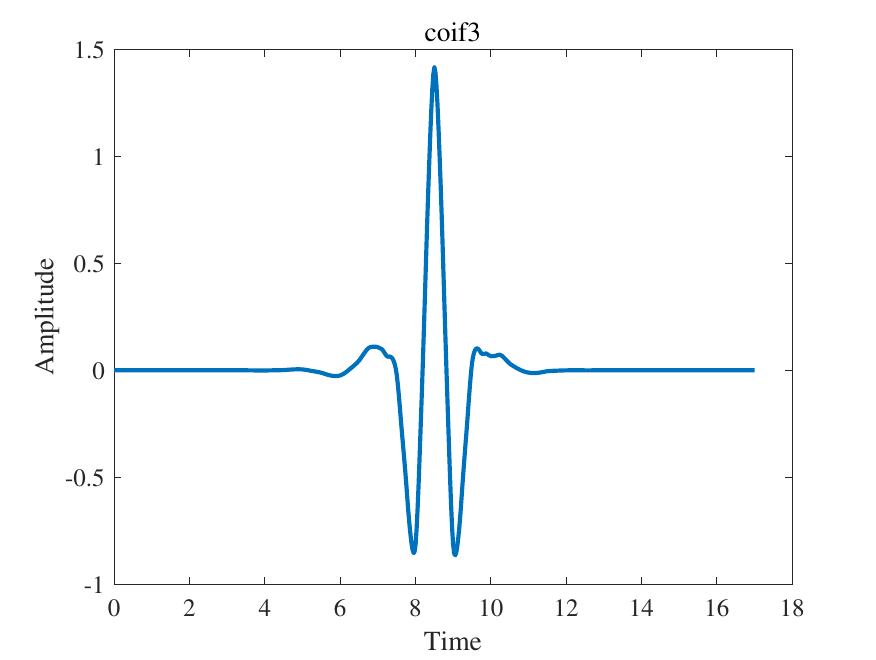}}
\subfigure[]{\label{fig:subfig2:rbio1.1}
\includegraphics[width=0.17\linewidth]{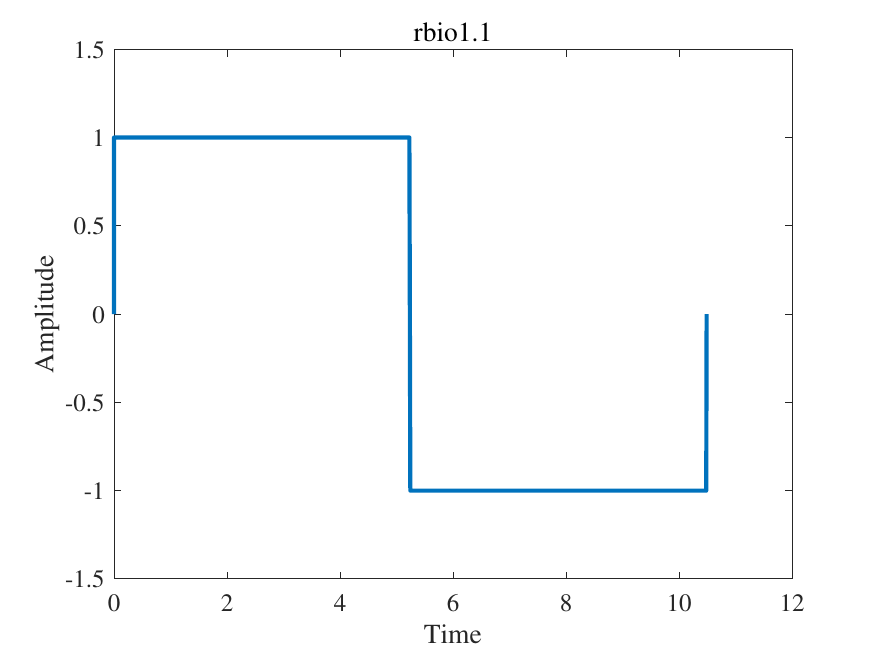}}
\centering
\caption{The waveforms of wavelet basis function. (a) haar; (b) db5; (c) sym5;  (d) coif3; (e) rbio1.1. }
\label{wave}
\end{figure*}

In order to make it easier to observe the features and attributes of the wavelet bases discussed previously, we plot time-domain waveforms for several commonly used wavelet bases, including haar, db5, sym5, coif3, and rbio1.1. These waveforms provide a clear depiction of the amplitude and shape of each wavelet as it changes over time in Fig. \ref{wave}. It can be seen that the coif3 wavelet is symmetrical, and the time-frequency waveforms of the haar and rbio1.1 wavelets are quite similar.

\begin{figure}[t]
    \centering
    \includegraphics[width=7.5cm]{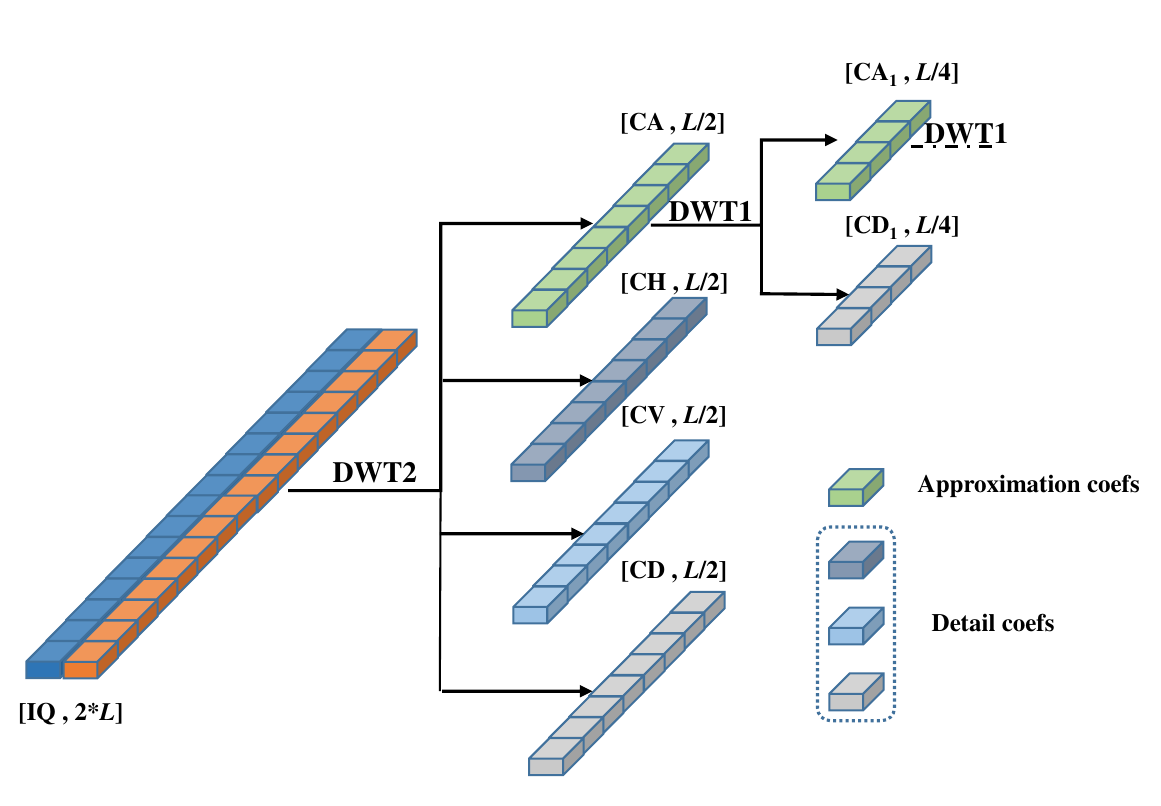}
    \caption{Hierarchy of DWT decomposition of IQ sequence.}
    \label{DWT}
\end{figure}

\subsubsection{Specific Decomposition Process} We propose a specific process for decomposing the received IQ sequence using DWT as shown in Fig. \ref{DWT}. In the process, we use two-dimensional DWT (DWT2) and one-dimensional DWT (DWT1) to decompose the signal. We first use DWT2 to decompose the IQ sequence to extract the two sets of coefficients, i.e., the approximate coefficients and detail coefficients of the signal. 
Given the scaling function $\varphi(x)$ and wavelet function $\omega(x)$ in advance, they can be combined into a two-dimensional scaling function and three two-dimensional wavelet functions as
\begin{equation}
\label{A}
\varphi(x,y)=\varphi(x)\varphi(y),
\end{equation}
\begin{equation}
\label{H}
\omega^H(x,y)=\omega(x)\varphi(y),
\end{equation}
\begin{equation}
\label{V}
\omega^V(x,y)=\varphi(x)\omega(y),
\end{equation}
\begin{equation}
\label{D}
\omega^D(x,y)=\omega(x)\omega(y).
\end{equation}
Therefore, we can decompose the IQ sequence $f(x,y)$ with a size of $2 \times L$ through DWT2 to obtain four components: the approximation coefficients $CA$ and the details in three orientations (horizontal $CH$, vertical $CV$, and diagonal $CD$), which are given as
\begin{equation}
\label{CA}
CA=\frac{1}{\sqrt{2 L}} \sum_{x=0}^{1} \sum_{y=0}^{L-1} f(x, y) \varphi{(x, y)},
\end{equation}
\begin{equation}
\label{CT}
CT=\frac{1}{\sqrt{2 L}} \sum_{x=0}^{1} \sum_{y=0}^{L-1} f(x, y) \omega^T{(x, y)}, T\in\{H,V,D\},
\end{equation}
where $H$, $V$, and $D$ represent horizontal, vertical, and diagonal orientations respectively. 
The approximate coefficients are the low-frequency part obtained after decomposition, which reflects the overall trend and general morphological characteristics of the signal. The detail parts are the high frequency parts obtained after decomposition, which reflect the detail information and local variation characteristics of the signal.

Since the dimensions of the four coefficients are reduced to $1 \times L/2$, we perform DWT1 decomposition of the approximate coefficient $CA$ as follows:
\begin{equation}
\label{CAi}
CA_i= \sum_{f=0}^{F-1} CA_{i-1}(2n-f)g[f],
\end{equation}
\begin{equation}
\label{CDi}
CD_i= \sum_{f=0}^{F-1} CA_{i-1}(2n-f)h[f],
\end{equation}
where $CA_i$ and $CD_i$ are the approximate coefficients and detail coefficients of the $i$-th order DWT1 respectively, $g[f]$ and $h[f]$ represent low pass filter and high pass filter respectively, $F$ is the length of filters. We define $E$ as the number of times to perform the DWT decomposition, including DWT2 and DWT1 decomposition. With this, we can obtain the coefficients of IQ sequences decomposed by wavelet transform, including 
an approximation coefficients set $CA_{E-1}$ and a detail coefficients set ${CB}=[CH,CV,CD,CD_1,\dots,CD_{E-1}]$ with $(E+2)$ detail coefficients by performing one DWT2 decomposition and $(E-1)$ times of DWT1 decomposition of the IQ sequence. In the introduction that follows, we will use the abbreviation ${DWT}(\cdot)$ to denote the decomposition process mentioned above. 
 
\subsection{Augmentation by Replacing}
A series of applications such as denoising, compression and feature extraction can be realized by analyzing and processing the detail coefficients. Based on the detail coefficients set ${CB}$ generated by the discrete wavelet transform of IQ sequence, we propose four data augmentation methods to obtain augmented samples by replacing the detail coefficients for improving the recognition accuracy of the CNNs network as discussed in detail below. We employ a sequential replacement approach to replace every detail coefficient within the detail coefficients set. Subsequently, we regrad the process of utilizing the replaced coefficients for reconstruction and resulting in the generation of $(E+2)$ novel samples as an augmentation operation. The number of augmentation operation is denoted as $D$.
\subsubsection{All Zero Sequence Replacing (AZSR)}

\begin{figure}[t]
    \centering
    \includegraphics[width=7cm]{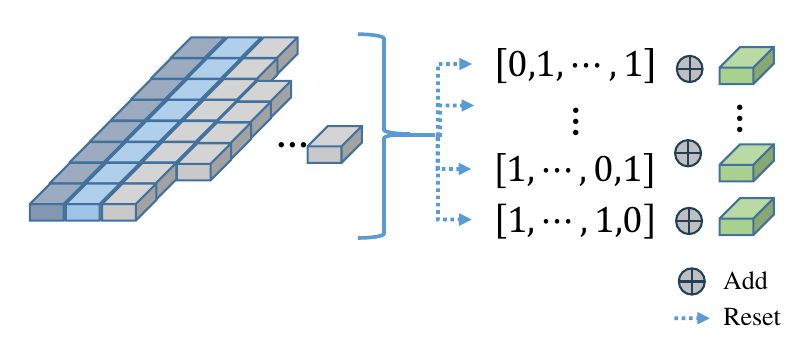}
    \caption{The process of data augmentation by AZSR method.}
    \label{AZSR}
\end{figure}

The AZSR augmentation method generates a sequence with all zero values whose length coincides with the length of any detail coefficient to be replaced in the detail coefficients set and then replaces the corresponding sequence in the detail coefficients set to obtain a new detail coefficient. After that, we use the replaced coefficients including the replaced detail coefficients and the approximation coefficients $CA_{E-1}$ for  reconstruction to obtain an augmented sample. The process of AZSR augmentation method is shown in Fig. \ref{AZSR}. Firstly, generate a vector $A_l$ with a length of $(E+2)$ and only $l$-th value of 0:
\begin{equation}
\label{AI}
A_l=[1,1,\dots,0,\dots,1,1], l=\left(0,1, \ldots, E+1\right).
\end{equation}
The all zero sequence replacing detail coefficients operation is performed by reseting the original decomposition detail coefficients set ${CB}$ with $CA_{E-1}$ is added to the replaced detail coefficients to obtain a new decomposition coefficient set $\mathcal{D}_l$
\begin{equation}
\label{CBi}
\mathcal{D}_l=A_l \odot {CB} \oplus CA_{E-1},
\end{equation}
where $\odot$ is the operation of reset which is to multiply the $l$-th sequence in the detail coefficients set by the $l$-th element in $A_l$, so that the values of the $l$-th detail coefficient in the detail coefficients set are all 0 and the remaining sequences in the detail coefficients set remain unchanged. $\oplus$ is the operation of adding coefficients.

\subsubsection{Random Zero Sequence Replacing (RZSR)}
\begin{figure}[t]
    \centering
    \includegraphics[width=6.5cm]{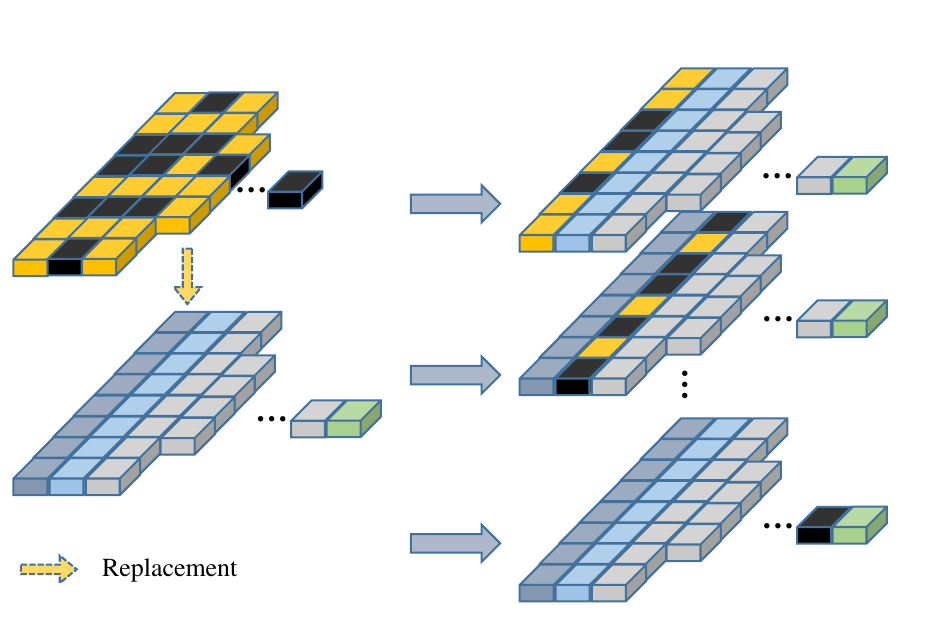}
    \caption{The process of data augmentation by RZSR method.}
    \label{RZSR}
\end{figure}

 The process of RZSR augmentation method is shown in Fig. \ref{RZSR}. 
The RZSR augmentation method is similar to the AZSR method in that both methods generate a sequence of the same length $L_0$ as the detail coefficient being replaced in the detail coefficients set. The difference is that RZSR generates a sequence with random zeros to replace the detail coefficient. The sequence used to generate replacement sequence can be represented as
\begin{small}
\begin{equation}
\label{zi}
 \hat{A_l}= \operatorname{choice}([0,1],L_0) ,
\end{equation}
\end{small}
where $\operatorname{choice}(Z,Y)$ represents randomly selecting elements from $Z$ to generate a sequence of length $Y$. To obtain the replacement sequence $B_l$, we multiply the generated vector $\hat{A_l}$ with the corresponding detail coefficient in the detail coefficients set to be replaced:
\begin{equation}
\label{CBz}
B_l=\hat{A_l}\otimes {CB_l},
\end{equation}
where $\otimes$ represents the operation of multiplying the corresponding elements of two vectors, $CB_{l}$ is the $l$-th detail coefficient in the detail coefficients set, $B_l$ is a sequences generated by replacing $CB_{l}$ with RZSR augmentation method, where any number of $B_l$ values are 0 and the remaining values remain unchanged.
By replacing $CB_{l}$ with ${B_l}$, a new coefficient set $\overline{\mathcal{D}_l}$ is obtained.
\begin{small}
\begin{equation}
\label{eq10}
   \begin{gathered}
\overline{\mathcal{D}_l}=[CA_{E-1},CH,CV, CD,CD_1,\dots, B_l,\dots, CD_{E-1}] .
\end{gathered}
\end{equation}
\end{small}

\subsubsection{Random Noise Sequence Replacing (RNSR)} 
Since the detail coefficients can be regarded as high-frequency noise, we can replace them with randomly generated noise sequences. In the RNSR augmentation method, in order to ensure power consistency, we make sure that the power of the generated noise sequence is consistent with the power of the detail coefficients to be replaced. We calculate the power of the replaced detail coefficient in the detail coefficients set as follows:
\begin{equation}
\label{power}
\beta= \sum_{l=0}^{L_0-1} CB_{l}^2,
\end{equation}
where $CB_{l}$ is the $l$-th detail coefficient in the detail coefficients. The generated noise sequence to replace $CB_{l}$ can be represented as
\begin{equation}
\label{B_L}
\hat{B_l}= \sqrt{\beta}\operatorname{randn}(1,L_0),
\end{equation}
where $\operatorname{randn}(p,q)$ function generates a sequence with a standard normal distribution and dimension $(p,q)$. Similarly, by replacing $CB_{l}$ with $\hat{B_l}$, a new coefficients set $\hat{\mathcal{D}_l}$ is obtained.

\subsubsection{Random Noise Sequence Replacing with Multiple Wavelets (RNSR-MW) }
The RNSR method initially just uses a single type of wavelet basis for decomposing IQ sequences. However, due to the wide range of wavelet basis options available, 
we can also use various wavelet basis functions to decompose the IQ sequences, and then replace the decomposed coefficients using the RNSR method. Afterwards, corresponding wavelet basis functions are used to reconstruct the replaced coefficients to obtain augmented samples. The above procedure is referred as the RNSR-MW augmentation method. 
Specifically, for a given signal $f$ and the modulation type $m$, we use different wavelet basis functions to decompose the corresponding IQ sequence to obtain the decomposition coefficients set
\begin{equation}
\label{dwt1}
B_{\omega_j}=DWT(f,\omega_j), j=(0,1,\dots,w-1),
\end{equation}
where $w$ is the number of wavelet basis functions we used. To obtain new decomposed coefficients set $\mathcal{D}_{\omega_{jl}}$, we replace the $l$-th detail coefficient in the detail coefficients set using the RNSR method. Therefore, by  replacing the $l$-th detail coefficient in the detail coefficients using $w$ types of wavelets, we can generate novel mixed replaced coefficients sets with $w$ wavelets, i.e., by replacing $l$-th detail coefficient in the detail coefficients set, we can expand the number of decomposition coefficients from one to $(w+1)$ and obtain the set $\mathcal{F}$:
\begin{equation}
\label{mix}
\mathcal{F}=\{(B_{\omega_{j}},m),(\mathcal{D}_{\omega_{0,l}},m),\dots,(\mathcal{D}_{\omega_{w-1,l}},m)\}.
\end{equation}
By considering the diversity of wavelet bases and combining decomposition coefficients from various wavelet bases at different time scales, this method aims to extract a greater number of signal features to improve recognition accuracy.

\subsection{IQ Sequence Reconstruction}
\begin{figure*}[t]
\centering
\subfigure[]{\label{fig:subfig2:CH}
\includegraphics[width=0.2\linewidth]{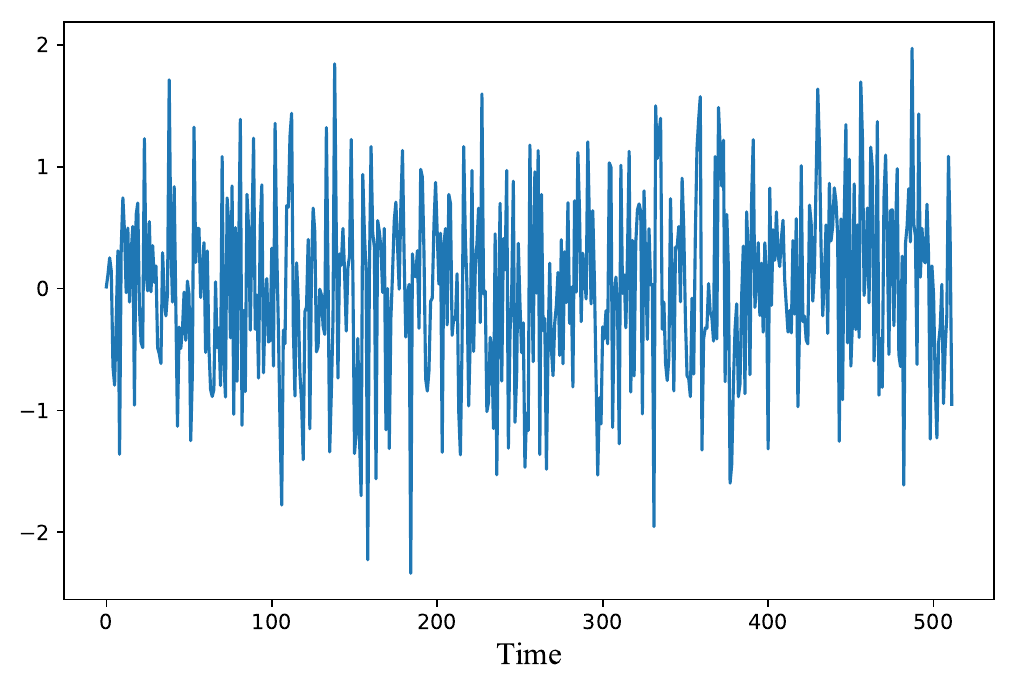}}
\hspace{5mm}
\subfigure[]{\label{fig:subfig2:CH_RZSR}
\includegraphics[width=0.2\linewidth]{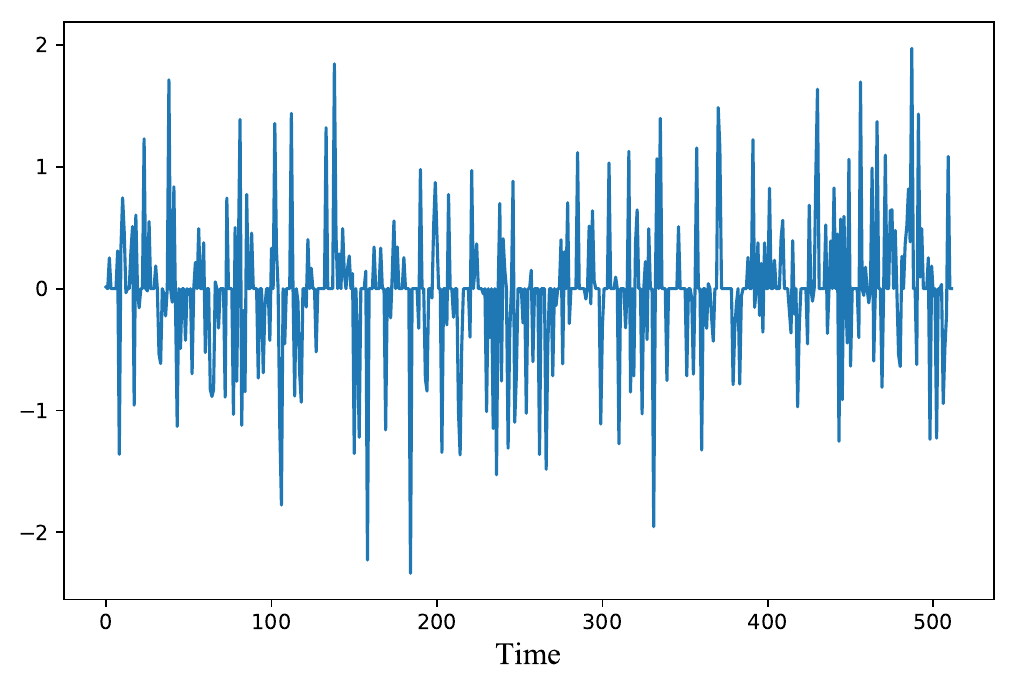}}
\hspace{5mm}
\subfigure[]{\label{fig:subfig2:CH_RNSR}
\includegraphics[width=0.2\linewidth]{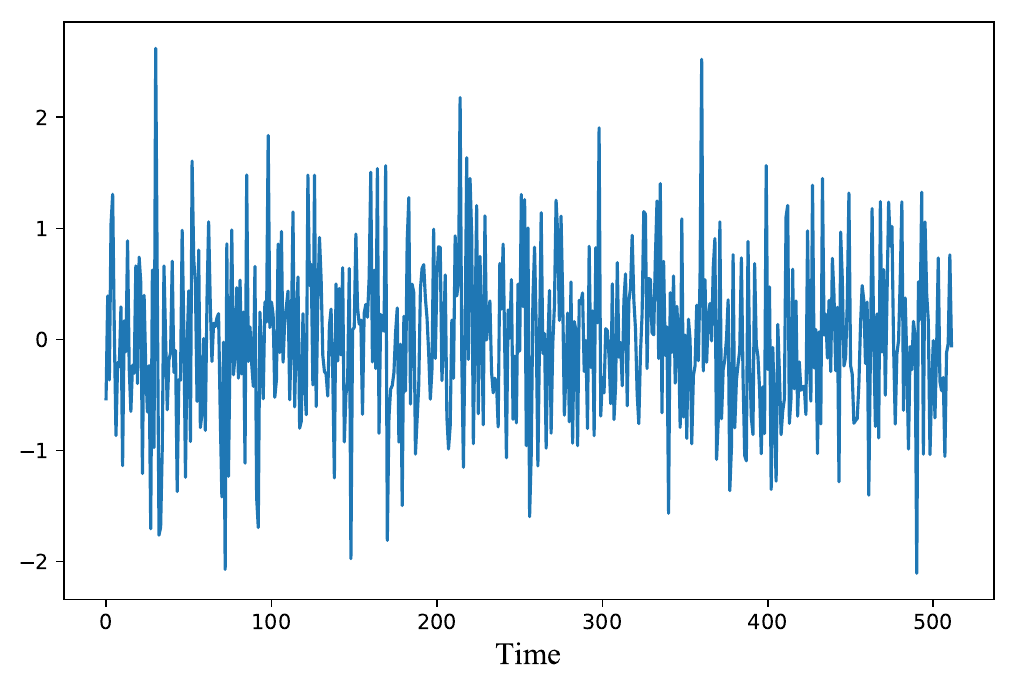}}
\centering
\caption{Replacing CH sequences with different replacement methods. (a) raw CH sequence; (b) RZSR; (c) RNSR.}
\label{ch2}
\end{figure*}

\begin{figure*}[t]
\centering
\subfigure[]{\label{fig:subfig3:IQ}
\includegraphics[width=0.23\linewidth]{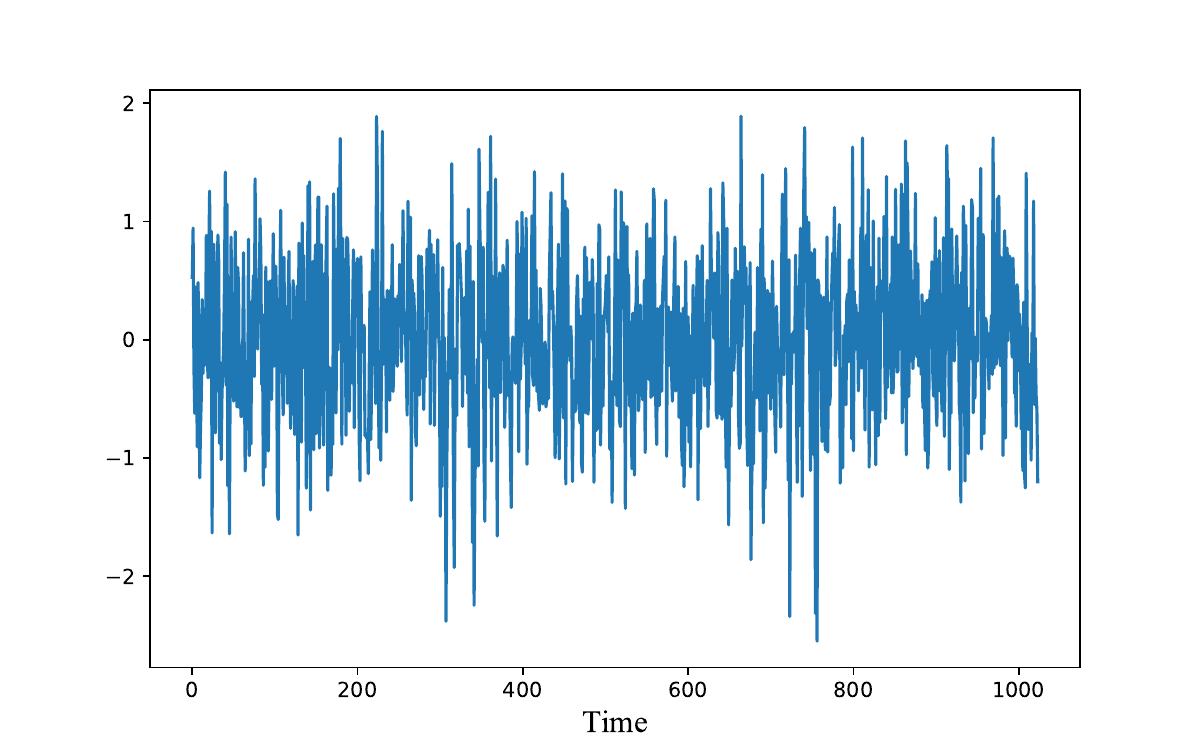}}
\subfigure[]{\label{fig:subfig3:AZSR}
\includegraphics[width=0.23\linewidth]{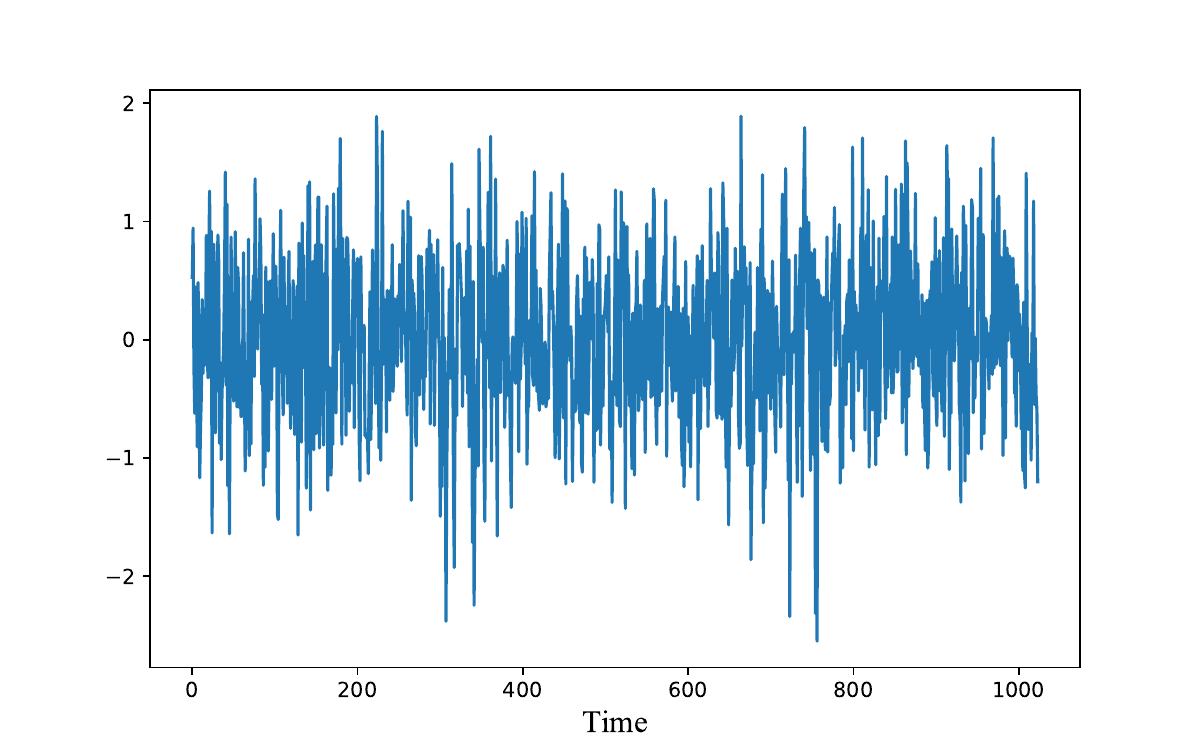}}
\subfigure[]{\label{fig:subfig3:RZSR}
\includegraphics[width=0.23\linewidth]{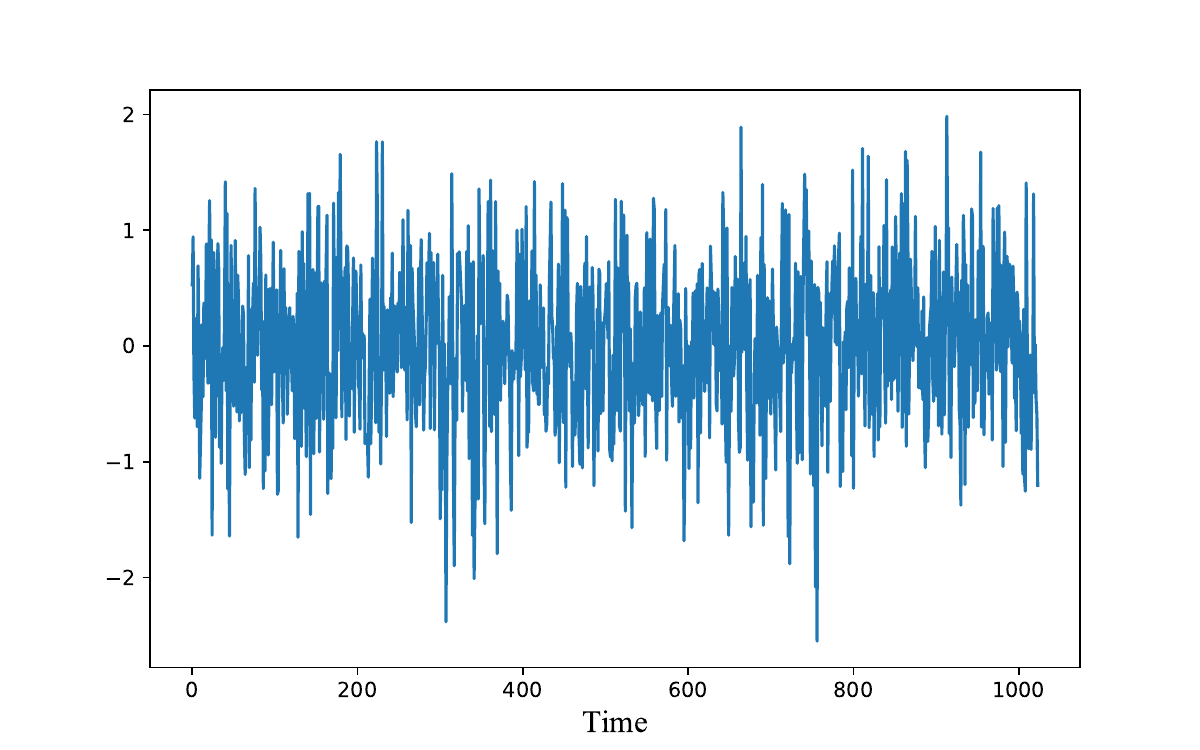}}
\subfigure[]{\label{fig:subfig3:RNSR}
\includegraphics[width=0.23\linewidth]{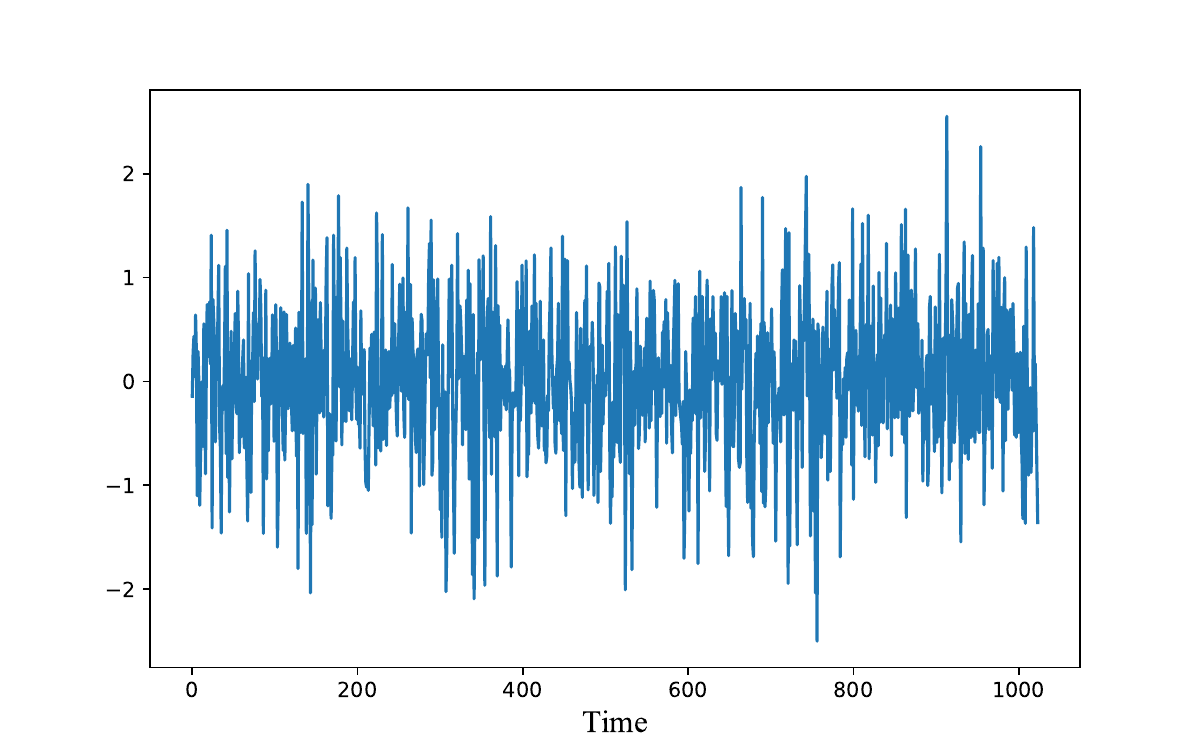}}
\centering

\caption{ I-channel sequences reconstructed by different replacement methods. (a) raw I-channel sequence; (b) AZSR; (c) RZSR; (d) RNSR.}
\label{RD}
\end{figure*}

By employing the aforementioned augmentation method, we obtain new decomposition coefficients set. Subsequently, we perform the reconstruction process by utilizing an approximate coefficients ste and the $(E+2)$ detail coefficients set. To begin with, we use the one-dimensional inverse discrete wavelet transform (IDWT1) to reconstruct the coefficients obtained from DWT1, thereby yielding the first-order approximate coefficients CA. The  expression for IDWT1 for each order of decomposition coefficients is as follows
\begin{equation}
\label{rec}
\small CA_{i-1}(n)=\sum_{f=0}^{F-1} g(n-2 f) CA_{i}(f)+\sum_{f=0}^{F-1} h(n-2 f) CD_{i}(f).
\end{equation}
Hence, the first order approximation coefficient CA is derived by reconstructing the approximation and detail coefficients acquired through the $(E-1)$ orders DWT1 using $(E-1)$ orders IDWT1  reconstruction. Subsequently, the reconstructed approximation coefficients and the three detail coefficients (CH, CV, and CD) obtained from one order DWT2 decomposition are employed in the two-dimensional IDWT (IDWT2) reconstruction process to yield the augmented sample $\widehat{IQ}(n)$. The expression for generating $\widehat{IQ}(n)$ samples using IDWT2 reconstruction can be expressed as
\begin{equation}
\label{IQ}
\widehat{IQ}(n)=C A \varphi(x, y)+C T \omega^T(x, y), T \in\{H, V, D\}.
\end{equation}

To acquire a more profound comprehension of the distinct replacement methods, we perform DTW2 using haar wavelet on IQ sequence with BPSK modulation type, and then use DTW1 for the approximate coefficients obtained from the decomposition. We then replace the CH in the decomposition detail coefficients using different methods. The results of the CH replacement are displayed in Fig. \ref{ch2}. As the RZSR replacement method results in all CH sequence values being zero, we only draw the original CH sequence along with the CH sequence replaced by the RZSR and RNSR method. The replaced coefficients are used to reconstruct new samples. We draw I-channel sequences generated by the reconstruction of different replacement methods in Fig. \ref{RD}. We can clearly observe that the I-channel sequences of the new sample reconstructed by the AZSR and RZSR replacement methods are slightly different from the original I-channel sequence, but the I-channel sequence of the new sample reconstructed by the RNSR replacement method is significantly different from the original I-channel sequence. This phenomenon shows that the RNSR method can better expand the characteristics of the signal to avoid the overfitting of the neural network in few-shot scenarios.

\section{Performance}

\subsection{Data Generation and Simulation Setting}

To verify the efficacy of our proposed augmentation method that leverages wavelet transform and deep learning, we use two datasets RML1024 \cite{9935275} and HKDD$\_$AMC36 \cite{10042021} through MATLAB2022a, details of which are shown in Table \ref{table2}. The simulation parameters used to generate the RML1024 dataset are nearly identical to those utilized for the HKDD$\_$AMC36 dataset generation. Hence, RML1024 dataset is similar to the HKDD$\_$AMC36 dataset, except that RML1024 contains 12 modulation types while HKDD$\_$AMC36 has an additional 24 modulation types, resulting in a total of 36 modulation types. For each of the datasets, the initial sequence of bits is randomly chosen from the binary values of 0 and 1. Each generated signal has a length of 1024, and an oversampling rate of 8 is applied, resulting in 128 symbols per signal. To shape the signal, the pulse-shaping filter employed is a root raised-cosine (RRC) filter with a truncated length of 6 symbols. The roll-off factor is chosen randomly from a range of 0.2 to 0.7. Additionally, the normalized frequency offset, ranging from $-$0.2 to 0.2, is randomly chosen with respect to the sampling frequency.
Both datasets have a signal-to-noise ratio (SNR) ranging from $-20$ dB to 30 dB, with increments of 2 dB. The ratio of training set to test set in both datasets is 1: 50. Furthermore, under each SNR, the number of each modulation type in the training set is 10. For RML1024 dataset, the number of training and test sets is 3120 and 156000, respectively. Similarly, for HKDD$\_$AMC36 dataset, the number of training and test sets is 9360 and 468000, respectively.
\begin{table*}[t]
\renewcommand\arraystretch{1.3}
\centering
\caption{The Dataset Information}
\label{table2}
\setlength{\tabcolsep}{2mm}{
\begin{tabular}{ccc}
\hline\hline
Dataset  & RML1024& HKDD$\_$AMC36\\ \hline
Data dimension    & 2*1024&2*1024 \\ \hline
Modulation types & \makecell[c]{BPSK, 8PSK,\\  2FSK,4FSK\\ 8FSK, QPSK,\\ OQPSK, 8PAM,\\ 4PAM, 64QAM, \\32QAM, 16QAM } & 
\makecell[c]{32PSK, 8FSK, 4FSK, 2FSK, 256QAM,\\ 16PSK, QPSK, OQPSK, 8PSK, BPSK,\\128QAM, 64QAM, 32QAM, 16QAM,\\ 256APSK, 128APSK, 64APSK, 32APSK, \\16APSK, 16PAM, 8PAM, 4PAM, OFDM-QPSK,\\ OFDM-BPSK, 8CPM, 4CPM, GMSK, OFDM-16QAM, \\ FM-MSK, AM-MSK, FM, AM, OOK, 8ASK, 4ASK, MSK}\\ \hline
Number of training set and test set   & 3120 , 156000 & 9360 , 468000 \\  \hline
SNR range  & \makecell[c]{[-20,30] dB,\\ 2 dB gap} &\makecell[c]{[-20,30] dB,\\ 2 dB gap}\\ \hline
Channel   & AWGN&AWGN \\ 
 \hline\hline
\end{tabular}}
\end{table*}

We conducted all simulations on a notebook with an Intel Core i9-12900HX CPU and NVIDIA RTX3080Ti GPU. Our simulations involve utilizing the PyTorch framework to recognize the modulation types of the data generated through simulation. During the training process, we set the maximum epoch to 50, with an initial learning rate of 0.07. The learning rate is decreased to 10\% of its previous value at epoch 20 and 40. The mini-batch size is set to 128 and increased to 256 when the number of samples rises significantly. Adam optimizer is used for training.

\subsection{Performance of Proposed Methods}
We define $E$ as the number of times to perform the discrete wavelet transform. Without any augmentation, the accuracy of the RML1024 dataset is 45.787\%, and the accuracy of the HKDD$\_$AMC36 dataset is 41.417\%. Under the same experimental conditions and parameters, we investigate the recognition accuracy of augmentation methods proposed by us on both datasets. In this scenario, the network we used is ResNet8 \cite{9935275}.
\subsubsection{AZSR}
\begin{table}[t]
\renewcommand\arraystretch{1.3}
\centering
\caption{The Accuracy of AZSR Augmentation Method }
\label{table_AZSR}
\setlength{\tabcolsep}{2.8mm}{
\begin{tabular}{c|cc}
\hline\hline
 \diagbox{E}{Dataset}&{RML1024} & {HKDD$\_$AMC36}\\ \hline
 0  & 45.787\%  & 41.417\%  \\ \hline
1  & 45.920\%  & 46.424\%  \\ \hline
2  & 45.979\%  & \pmb{47.676\%}  \\ \hline
3  & \pmb{48.099\%}  & 47.172\% \\ \hline
4  & 45.131\%  & 46.772\%  \\ \hline
5  & 44.951\%  & 46.526\%  \\ 
\hline\hline
\end{tabular}}
\end{table}
The AZSR augmentation method involves replacing the sequence in the detail coefficients with a newly generated sequence that is of the same length as the sequence to be replaced in the detail coefficients, and has all its values set to 0. We conduct the discrete wavelet transform on the IQ sequence up to 5 levels, i.e., $E =$ 1, 2, 3, 4, and 5. The used wavelet is haar and the simulation results are show in Table \ref{table_AZSR}. Based on the experimental results, we observe that increasing the number of performing the discrete wavelet transform on the IQ sequence initially leads to an improvement in accuracy on both datasets, followed by a decrease in accuracy. The improvement of performance on the RML1024 dataset is not as significant as that on the HKDD$\_$AMC36 dataset. For instance, the maximum performance gain achieved on the RML1024 dataset is 2.3\% with $E =$ 3, while the performance gain on the HKDD$\_$AMC36 dataset is 6.3\% with $E =$ 2. Meanwhile, on the RML1024 dataset, the augmented method performs worse than that of the method without augmentation when the number of performing wavelet transform is large ($E =$ 4, 5). On the HKDD$\_$AMC36 dataset, even with the highest number of decomposition level ($E =$ 5), the accuracy of the augmented method remains superior to that of the method without augmentation.

\subsubsection{RZSR}
\begin{table}[t]
\renewcommand\arraystretch{1.3}
\centering
\caption{The Accuracy of RZSR Augmentation Method }
\label{table_RZSR}
\setlength{\tabcolsep}{1.2mm}{
\begin{tabular}{c|ccc|ccc}\hline\hline
{Dataset}&\multicolumn{3}{c|}{RML1024} &\multicolumn{3}{c}{HKDD$\_$AMC36}\\ \hline 
 \diagbox{D}{E}& 1&2&5 & 1&2&5 \\ \hline 
1  & 45.565\%  & 45.069\% & 45.089\% & 44.822\%  & 44.830\% & 46.892\% \\ \hline
2  & 45.332\%  & 45.710\% &\ 47.917\% & 45.687\%  & 46.980\% &\pmb{47.221\%} \\ \hline
4  & 46.468\%  & 46.530\% &49.032\% & 45.526\%  & 47.195\% &46.965\% \\ \hline
6 & 47.664\%  & 47.194\% &48.980\% & 45.722\%  & 46.438\% &46.289\% \\ \hline
8  & 49.350\%  & 48.016\% &\pmb{49.860\%}& 45.351\%  & 46.585\% &46.338\% \\ \hline\hline
\end{tabular}}
\end{table}
The RZSR augmentation method is similar to the AZSR method, where the elements of the replaced detail coefficients are randomly set to 0 instead of forcibly being replaced by 0. We consider the effect of the number of augmentation operation and the number of discrete wavelet transform iterations on the performance of the RZSR method. Specifically, the wavelet we used is haar and we set $D$ to be 1, 2, 4, 6, and 8, and $E$ to be 1, 2, and 5. Results are shown in Table \ref{table_RZSR}.
 It is apparent that the RZSR-based augmentation method has a performance improvement compared to the method without augmentation on the RML1024 dataset when $D$ is greater than or equal to 2 and $E$ is greater than or equal to 2.
 At $E =$ 5 and $D =$ 8, the maximum recognition accuracy is 49.806\%, which has a 4\% performance gain compared to the method without augmentation. In the case of the HKDD\_AMC36 dataset, the RZSR-based method we proposed consistently outperformes the method without augmentation for all values of $D$ and $E$. The highest recognition accuracy is achieved when $D = 2$ and $E = 5$ with a performance gain of 6\% compared to methods without augmentation.



\subsubsection{RNSR}
The RNSR augmentation method is designed to maintain the power of the original sequence in the detail coefficients. Specifically, it generates a random noise sequence with the same power as the sequence to be replaced in the detail coefficients. This helps to preserve the original characteristics of the signal while augmenting the dataset. The experimental results in the Table \ref{table_AZSR} show that the AZSR augmentation method performs well on both datasets in the three orders discrete wavelet transform iterations. Thus, in the following experiments, we fix the number of performing discrete wavelet transform to 3 for both datasets. Similarly, the haar wavelet is used as the wavelet basis and we set $D$ to be 1, 2, 4, 6, and 8. Table \ref{table_RNSR} displays the simulation results.

\begin{table}[t]
\renewcommand\arraystretch{1.3}
\centering
\caption{The Accuracy of RNSR Augmentation Method}
\label{table_RNSR}
\setlength{\tabcolsep}{2.8mm}{
\begin{tabular}{c|cc}
\hline\hline
 \diagbox{D}{Dataset}&{RML1024} & {HKDD$\_$AMC36}\\ \hline
1  & 46.766\%  & 48.671\%  \\ \hline
2  & 51.240\%  & 50.007\%  \\ \hline
4  & 56.754\%  & 50.669\% \\ \hline
6  & \pmb{57.308\%}  &\pmb{50.707\% } \\ \hline
8  & 57.307\%  & 50.619\%  \\ 
\hline\hline
\end{tabular}}
\end{table}

According to the experimental results, the RNSR augmentation method achieves the highest recognition accuracy on both datasets when the number of discrete wavelet transform iterations is 6. Specifically, the RNSR augmentation method outperforms the previous two methods, i.e., AZSR and RZSR augmentation methods. In the RML1024 dataset, the highest accuracy achieved by AZSR and RZSR augmentation methods is 48.099\% and 49.860\%, respectively, while the RNSR augmentation method achieves the highest accuracy of 57.308\%, which is about 8\% higher. Similarly, in the HKDD\_AMC36 dataset, the RNSR augmentation method achieves the highest accuracy of 50.707\%, while the highest accuracy achieved by AZSR and RZSR augmentation methods is 47.676\% and 47.221\%, respectively, with a gap of about 3\%. 
\begin{figure}
\centering
\subfigure[]{\label{RNSR_SNR:RNSR_RML1024}
\includegraphics[width=0.8\linewidth]{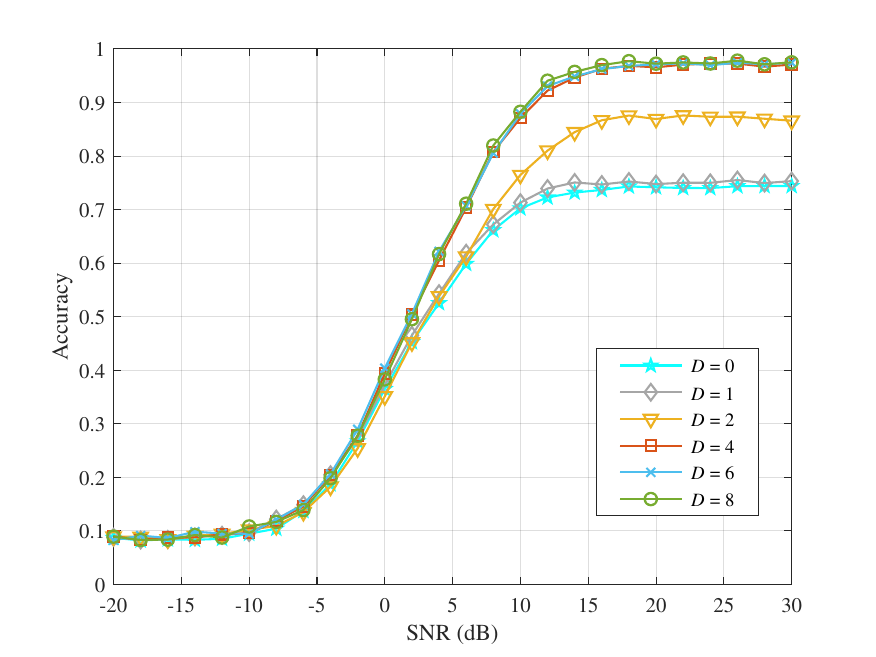}}
\subfigure[]{\label{fig1:subfig:RNSR_HKDD}
\includegraphics[width=0.8\linewidth]{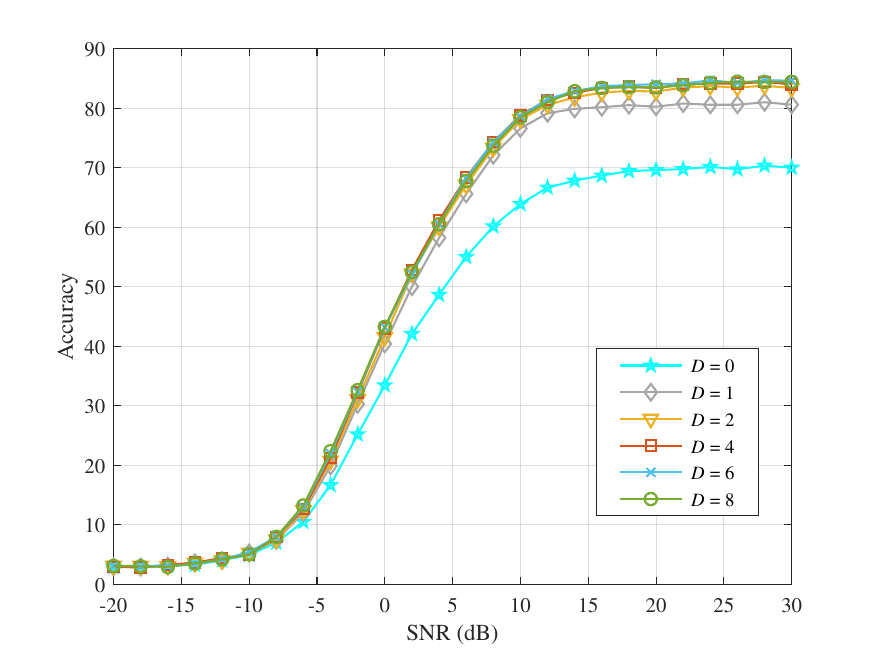}}
\centering

\caption{Recognition accuracy of RNSR method on the two datasets. (a) RML1024 and (b) HKDD$\_$AMC36.}
\label{RNSR_SNR}
\end{figure}

\begin{figure*}
\centering
\subfigure[]{\label{db1_4:IQ_hunxiao}
\includegraphics[width=0.32\linewidth]{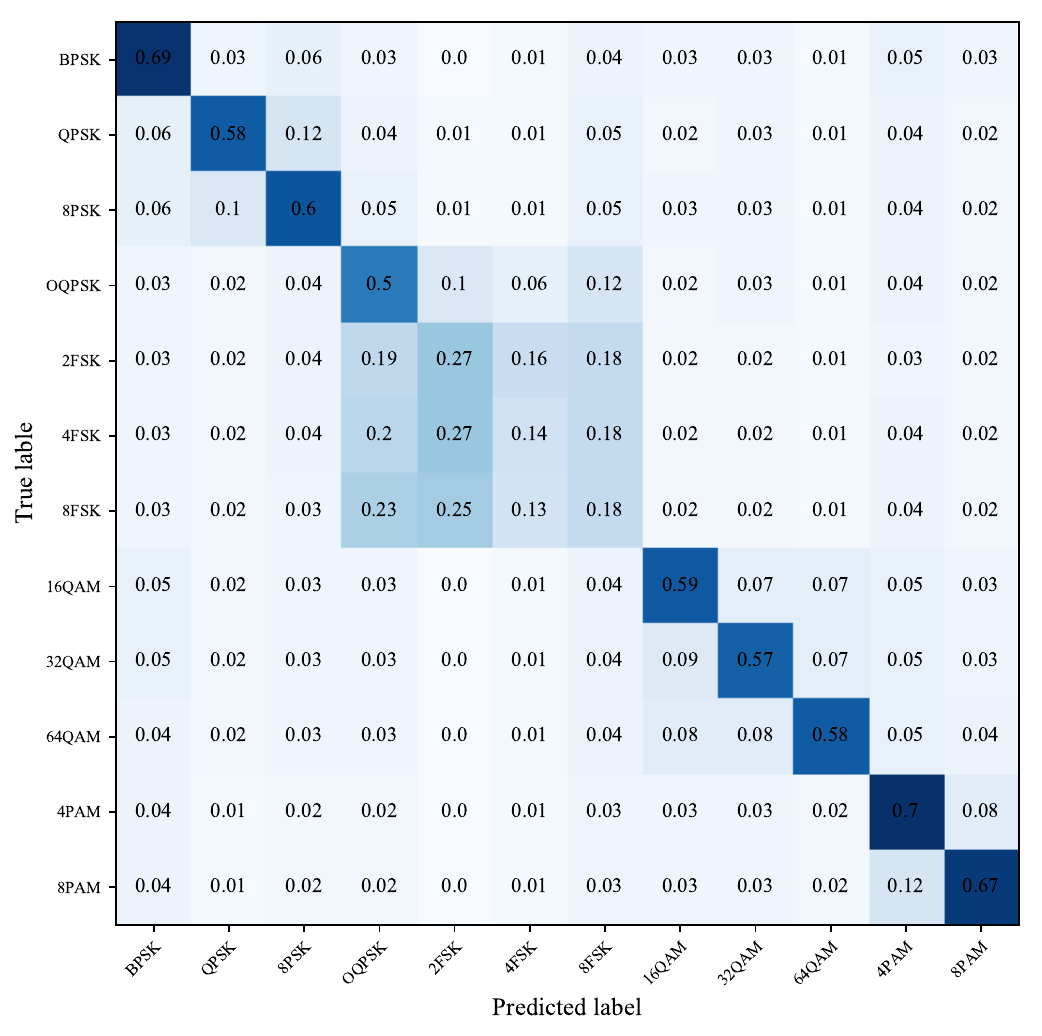}}
\subfigure[]{\label{fig:subfig1:db1_1}
\includegraphics[width=0.32\linewidth]{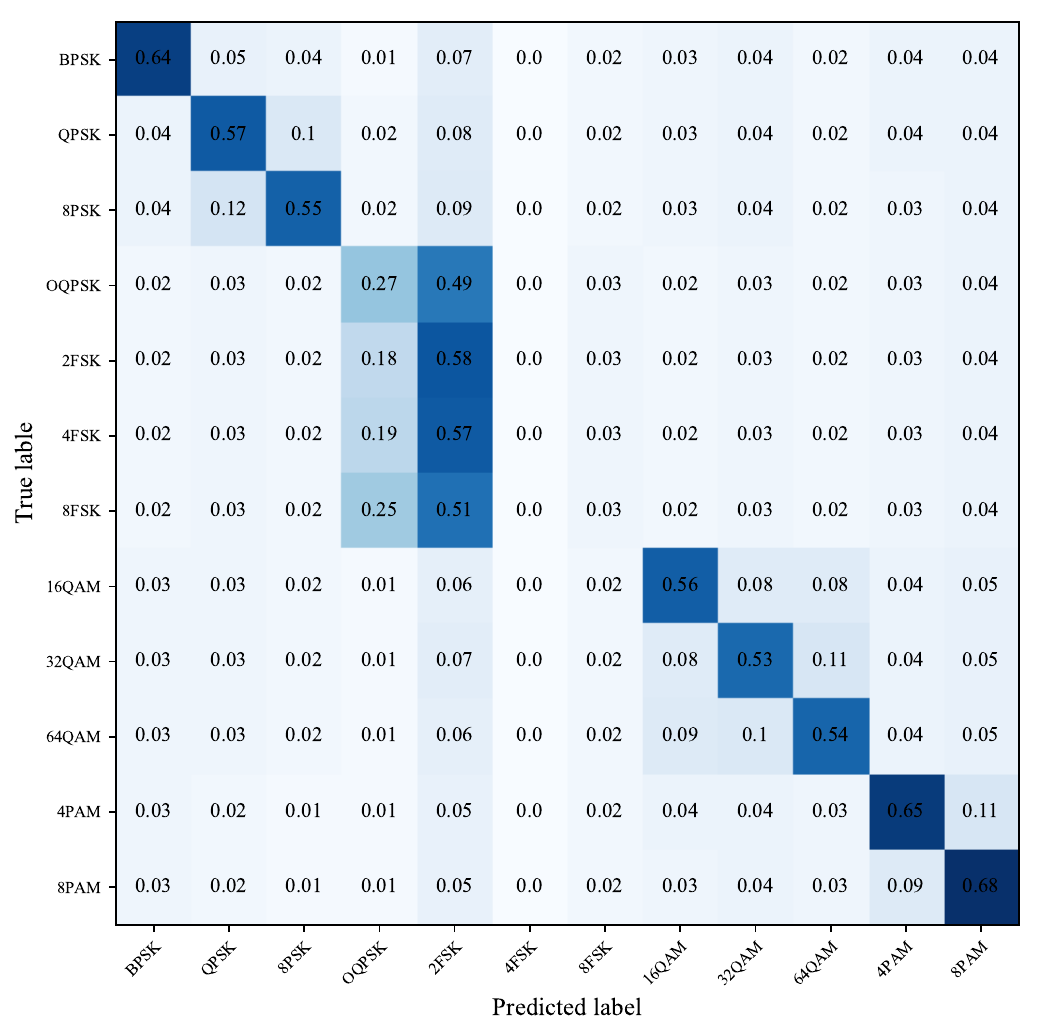}}
\subfigure[]{\label{fig:subfig1:db1_2}
\includegraphics[width=0.32\linewidth]{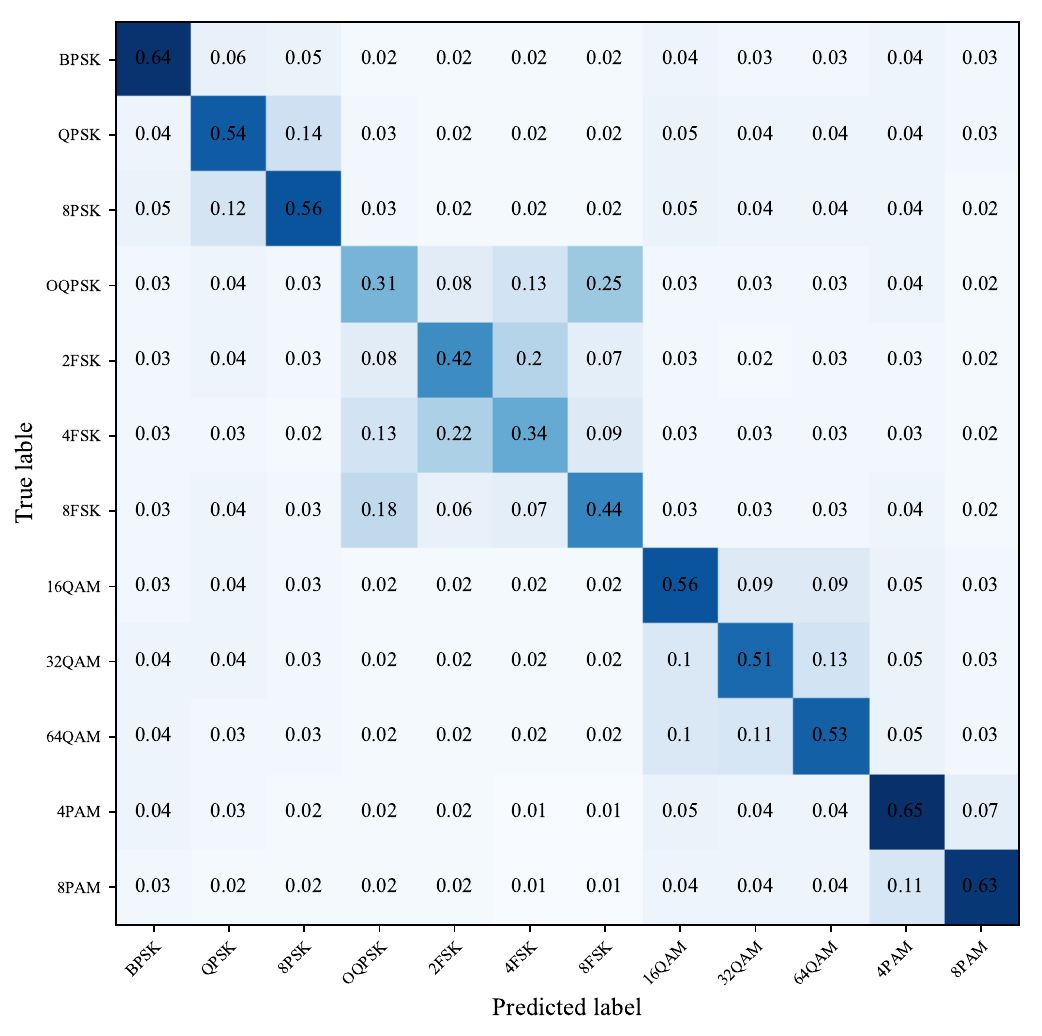}}
	\qquad
\subfigure[]{\label{fig:subfig2:db1_4}
\includegraphics[width=0.32\linewidth]{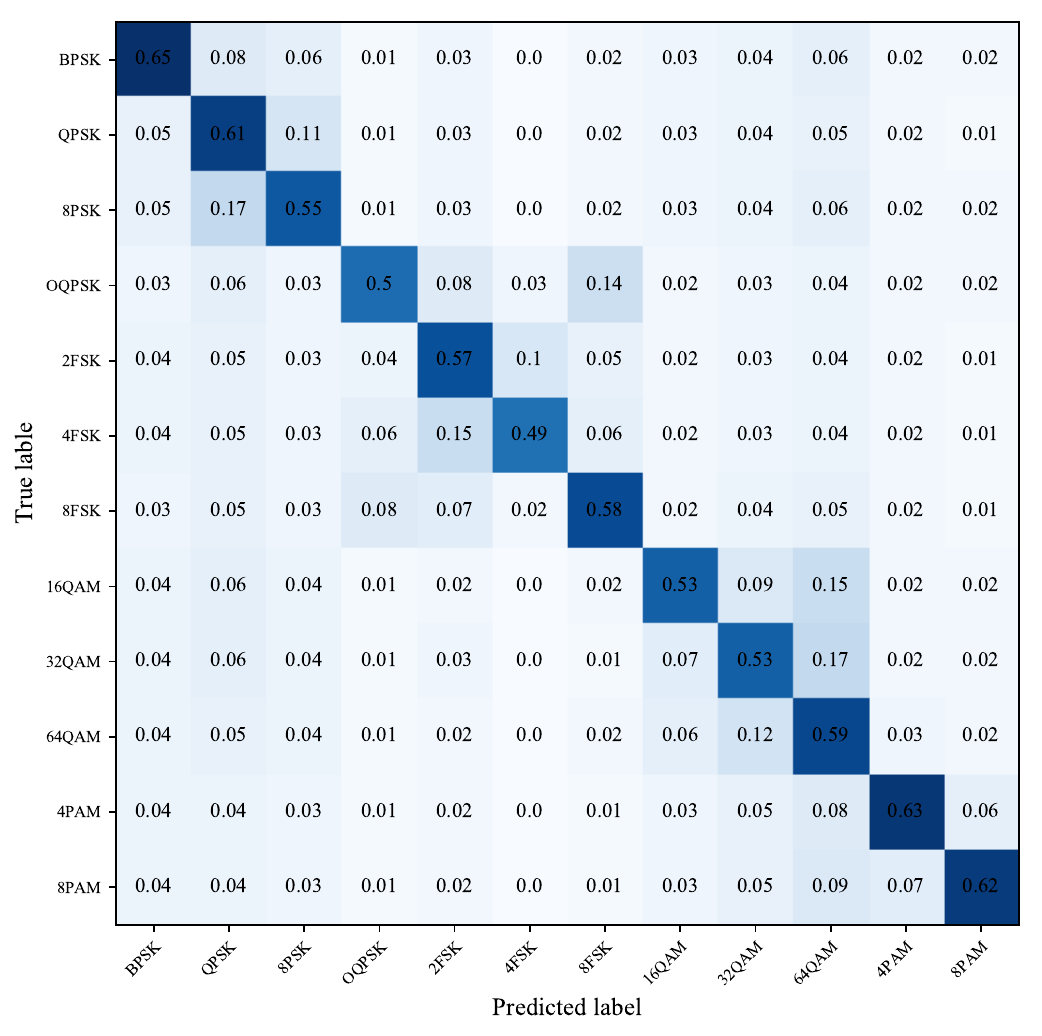}}
\subfigure[]{\label{fig:subfig2:db1_6}
\includegraphics[width=0.32\linewidth]{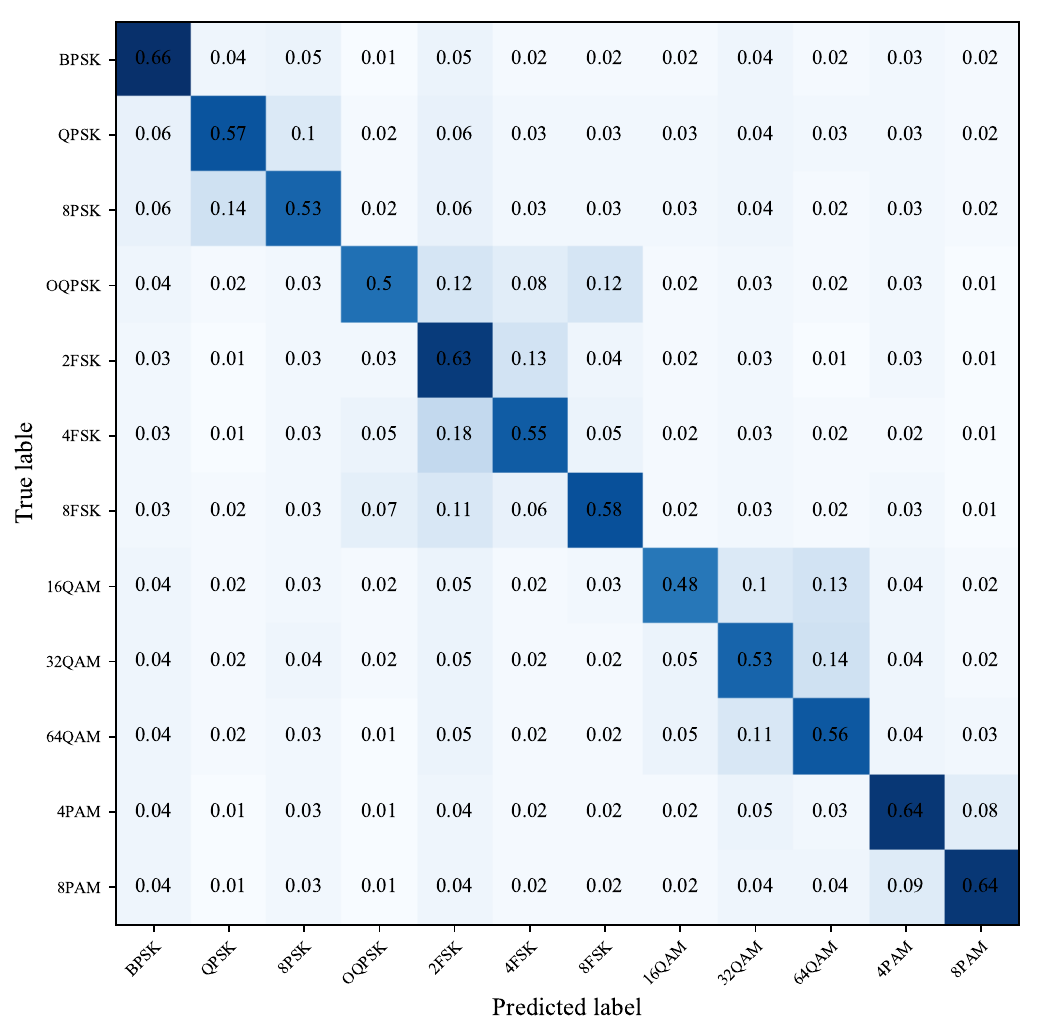}}
\subfigure[]{\label{fig:subfig2:db1_8}
\includegraphics[width=0.32\linewidth]{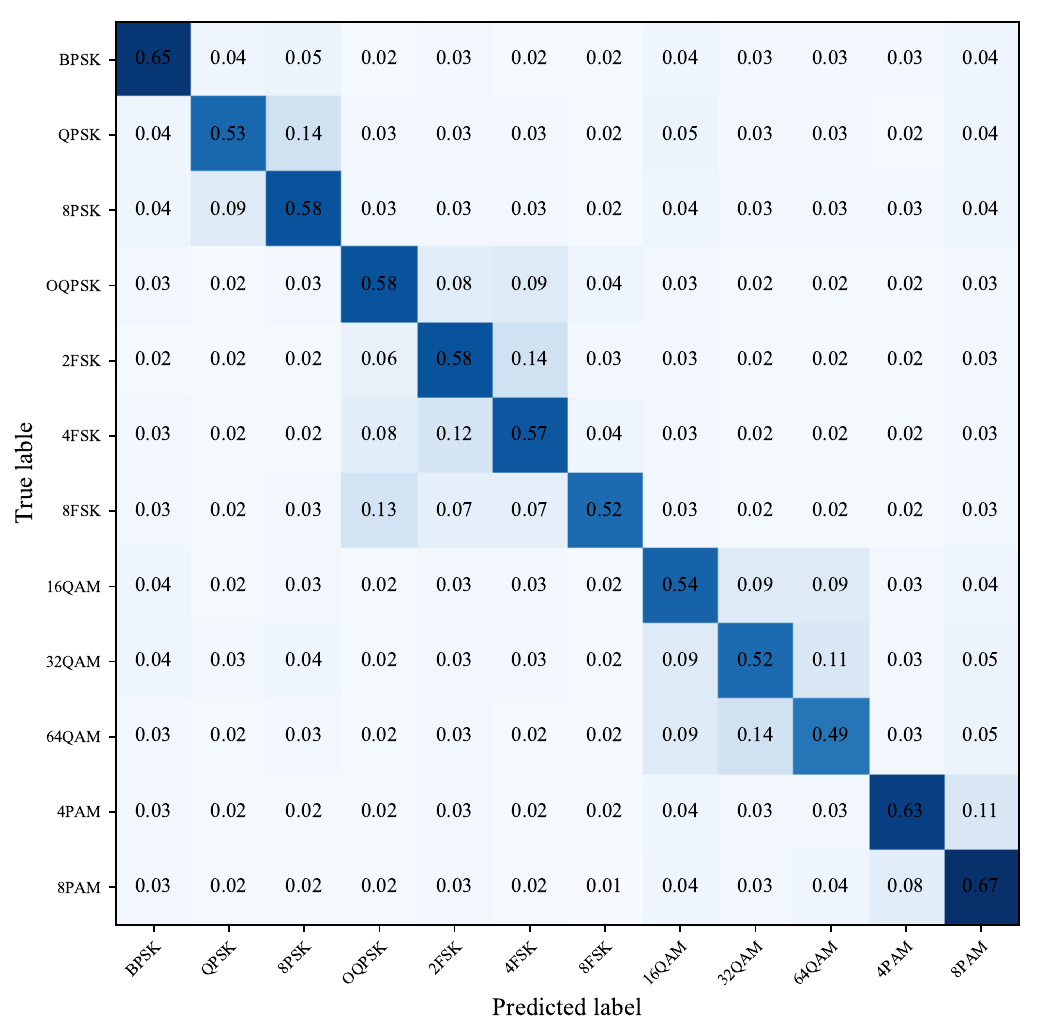}}
\centering
\caption{The confusion matrices under the different number of augmentation operation on the RML1024 dataset. (a) $D =$ 0; (b) $D =$ 1; (c) $D =$ 2; (d) $D =$ 4; (e) $D =$ 6; (f) $D =$ 8 .}
\label{db1_4}
\end{figure*}

We also plot the accuracy of RNSR augmentation method on two datasets at different SNR in Fig. \ref{RNSR_SNR}. The results clearly demonstrate that our proposed RNSR augmentation method achieves higher accuracy compared to the method without augmentation on both the RML1024 and HKDD\_AMC36 datasets. On the RML1024 dataset, the improvement is observed across the SNR range of 5 dB to 30 dB, with similar recognition accuracy at $D =$ 4, 6, 8. On the HKDD\_AMC36 dataset, the improvement is observed in the wide SNR range from $-$3 dB to 30 dB, with a performance gain that increases as the SNR increases and the recognition accuracy is roughly the same for $D =$ 2, 4, 6, 8. To visualize the recognition accuracy of each modulation type, we draw confusion matrices under the different number of augmentation operation on the RML1024 dataset in Fig. \ref{db1_4}. From the confusion matrix, we can see the classification accuracy of each modulation type and infer in which modulation types can get greater improvement. It also demonstrates that RNSR augmentation method has a notable improvement in recognition accuracy for the FSK category modulation types, including 2FSK, 4FSK, and 8FSK. Specifically, the improvement in accuracy for these modulation types is approximately 30\%, which is a considerable improvement.

\subsubsection{RNSR-MW}

To take into account the variability of different wavelet bases and the local features that are extracted from different wavelet bases at different time scales, we apply the augmentation RNSR method on IQ sequences using different wavelet base functions. Specifically, we use the haar, db5, sym5, coif3, and rabio1.1 wavelet bases to perform the decomposition and reconstruction. After decomposing and reconstructing using different wavelet bases to generated new samples, we combine the generated new samples by mixing them, and used these mixed samples for modulation recognition. By doing so, we can leverage the diversity and local features of different wavelet bases to enhance the discriminative power of the generated samples and improve the performance of the modulation recognition task. The dataset we used in this scenario is RML1024 and the number of discrete wavelet transform iterations is 3. The recognition accuracy is presented in Table \ref{table_RNSR} and Table \ref{mix_wave}.
\begin{table}[t]
\renewcommand\arraystretch{1.3}
\centering
\caption{The Accuracy of Different Wavelet Bases and RNSR-MW Method}
\label{mix_wave}
\setlength{\tabcolsep}{1mm}{
\begin{tabular}{c|ccccc}
\hline\hline
 \diagbox{D}{Wavelet} &db5&sym5&coif3&rbio1.1&mixing\\ \hline
1    & 46.560\% &46.735\% &46.690\% &46.662\% &\pmb{60.574\%} \\ \hline 
2   & 54.461\% &51.480\% &56.995\% &52.252\% &\pmb{61.328\%} \\ \hline 
4   & 57.212\% &58.409\% &60.301\% &56.177\%  &\pmb{60.722\% }\\ \hline  
6   & 60.834\% &58.930\% &60.432\% &56.317\% &\pmb{60.876\%} \\ \hline  
8   & \pmb{60.629\%} &57.142\% &60.442\% &58.074\% &59.918\% \\  
\hline\hline
\end{tabular}}
\end{table}

\begin{figure}[t]
    \centering
    \includegraphics[width=7.2cm]{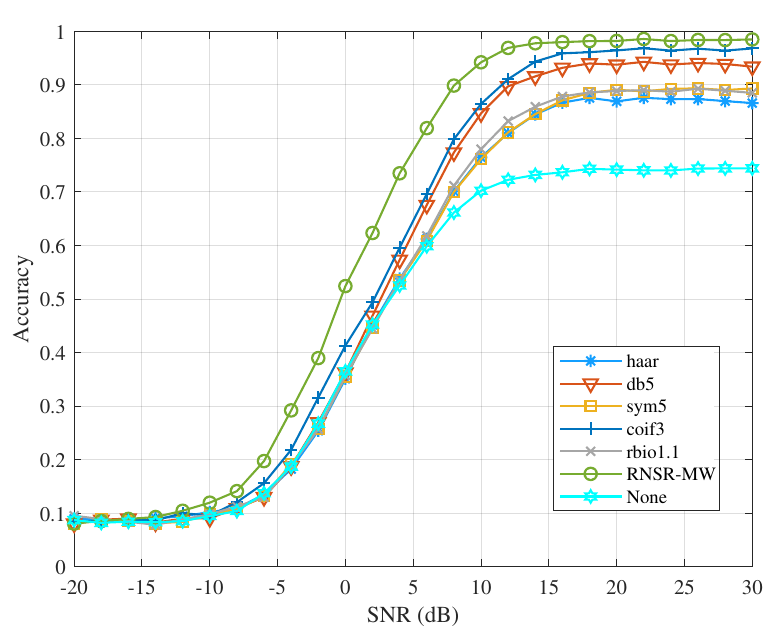}
    \caption{The accuracy of different wavelet bases and RNSR-MW method at each SNR when $D$ = 2 on RML1024 dataset.}
    \label{SNR_mwRZSR}
\end{figure}

It is evident that the RNSR-based method achieves notably higher modulation recognition accuracy when employing the coif3 and db5 wavelet bases compared to other wavelet bases when the number of augmentation operation ranges from 1 to 8. This suggests that coif3 and db5 wavelet bases are more effective in extracting signal sequence features at different time scales, leading to better performance in modulation recognition. Our proposed RNSR-MW augmentation method has shown improvements in the number of augmentation operation (from 1 to 6) compared to using a single wavelet base of RNSR-based method. With $D =$ 8, RNSR-based method generates an excessive number of augmented samples using a single wavelet. As a consequence, the network overfits during training by RNSR-MW method, resulting in a decrease in  accuracy. The performance gain is more significant when the number of generating replacement sequences is smaller. For instance, at $D =$ 1, the performance gain is as high as 14\% compared to the RNSR-based method that uses a single wavelet base. At $D =$ 6, the performance gain is slightly higher than that of the RNSR-based method using db5 and coif3 wavelets. We also plot their accuracy at different SNR in Fig. \ref{SNR_mwRZSR}. The results indicate that the RNSR-based augmentation method yields improved performance in the SNR range of 5 dB to 30 dB.
The recognition accuracy of our proposed RNSR-MW method outperforms that of using the single wavelet base RNSR-based augmentation method when the SNR is within the range of -12 dB to 30 dB.

\begin{figure*}
\centering
\subfigure[]{\label{fig:subfig2:coif3_hunxiao}
\includegraphics[width=0.32\linewidth]{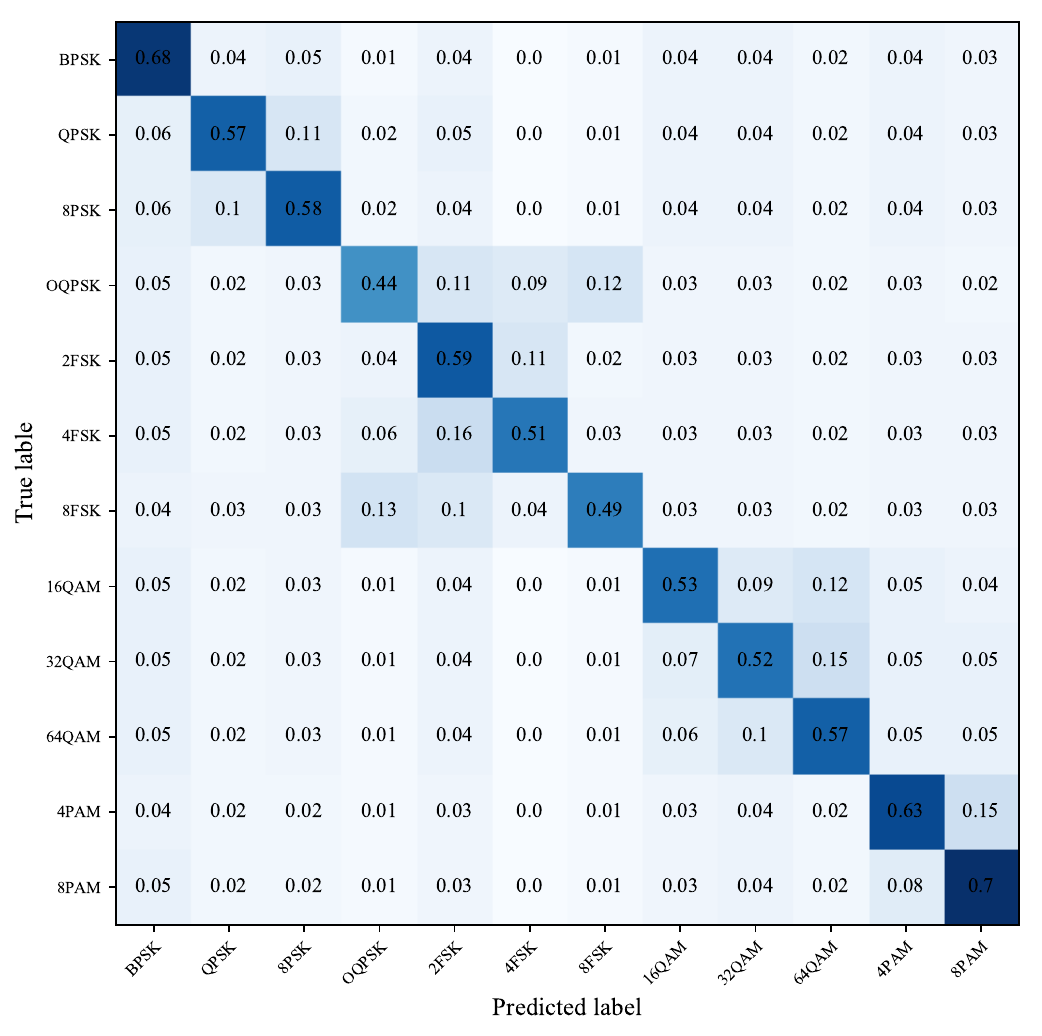}}
\subfigure[]{\label{fig:subfig2:rbio1.1_hunxiao}
\includegraphics[width=0.32\linewidth]{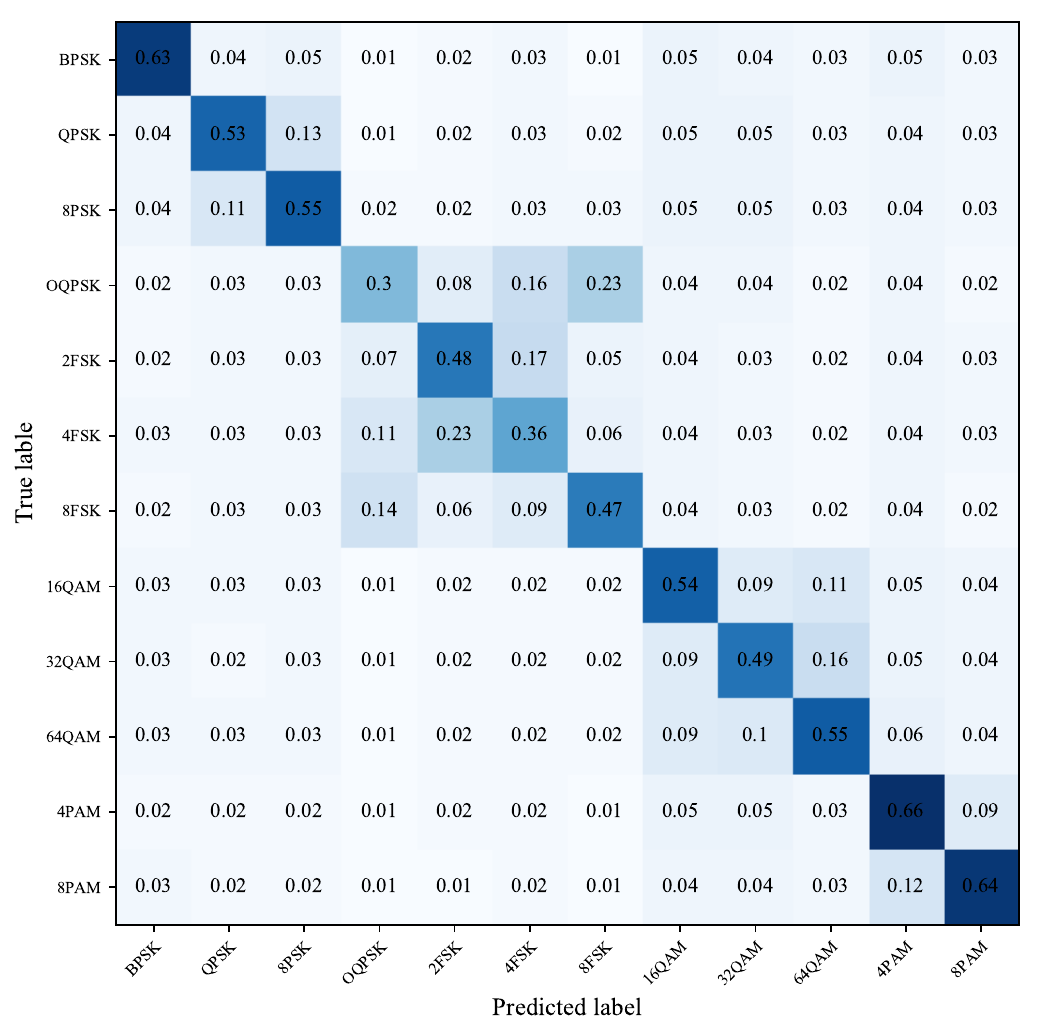}}
\subfigure[]{\label{fig:subfig2:mixing_hunxiao}
\includegraphics[width=0.32\linewidth]{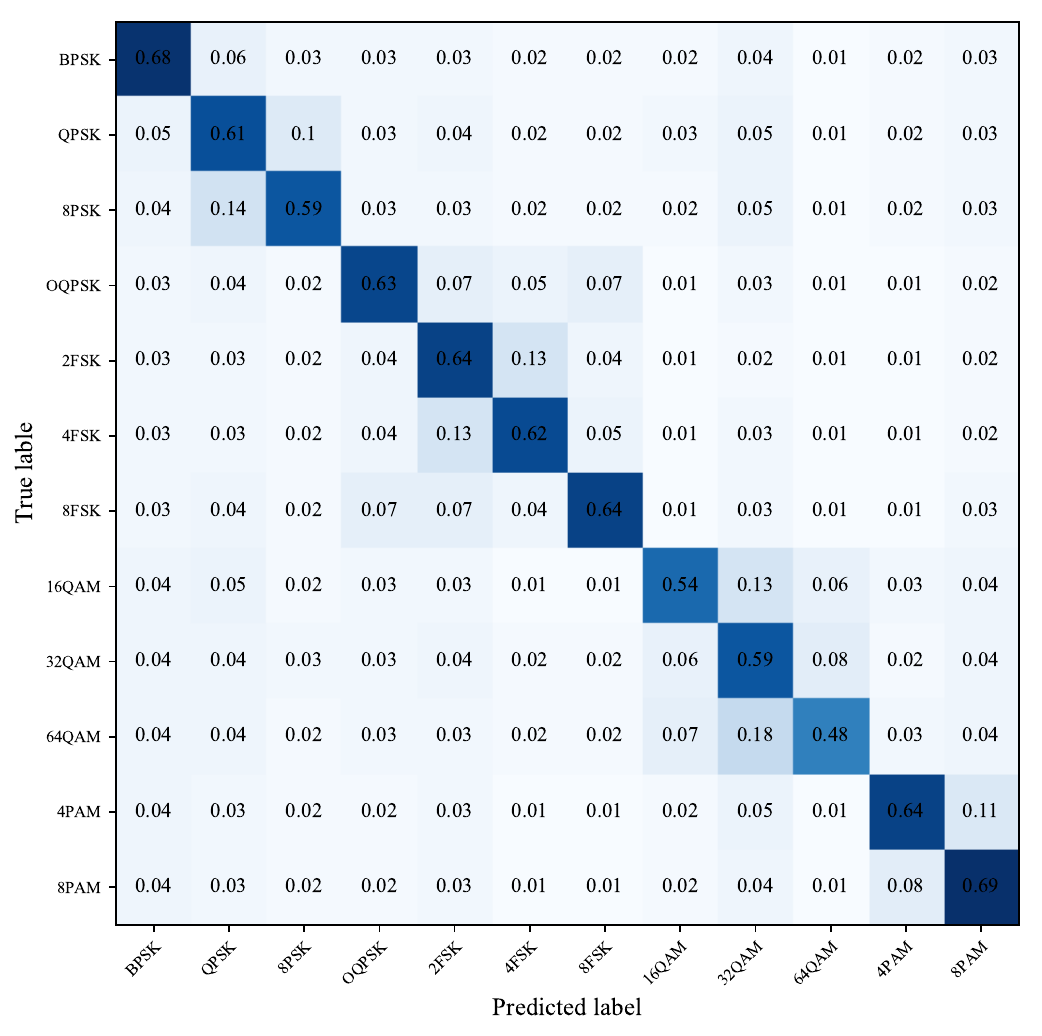}}
\centering
\caption{The confusion matrices of different wavelet bases under RNSR-MW method. (a) haar; (b) db5; (c) sym5; (d) coif3; (e) rbio1.1; (f) RNSR-MW.}
\label{input1}
\end{figure*}

Moreover, we present the confusion matrices of these methods in Figure \ref{input1}. Based on the Fig. \ref{input1} and Fig. \ref{db1_4:IQ_hunxiao}, it can be concluded that the RNSR-MW we proposed exhibits significant improvements in recognition accuracy compared to the method without augmentation in OQPSK, 2FSK, 4FSK, and 8FSK modulation types, with improvements of 13\%, 37\%, 48\%, and 46\%, respectively. It is worth noting that RNSR-MW outperforms RNSR using the best-performing coif3 wavelet in several modulation types, including OQPSK, 4FSK, and 8FSK, with improvements of 9\%, 11\%, and 15\%, respectively.

\begin{figure*}
\centering
\subfigure[]{\label{fig:subfig1:db1_hunxiao}
\includegraphics[width=0.32\linewidth]{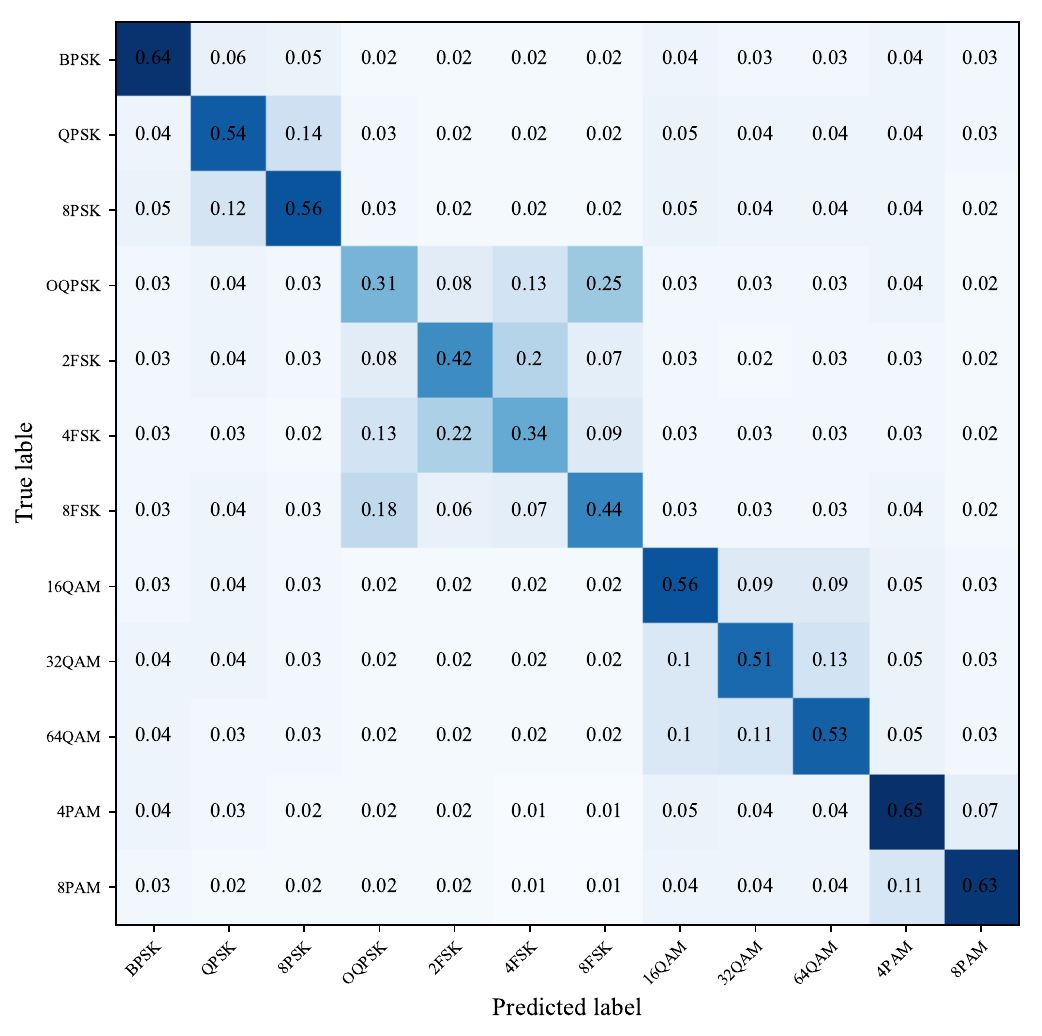}}
\subfigure[]{\label{fig:subfig1:db5_hunxiao}
\includegraphics[width=0.32\linewidth]{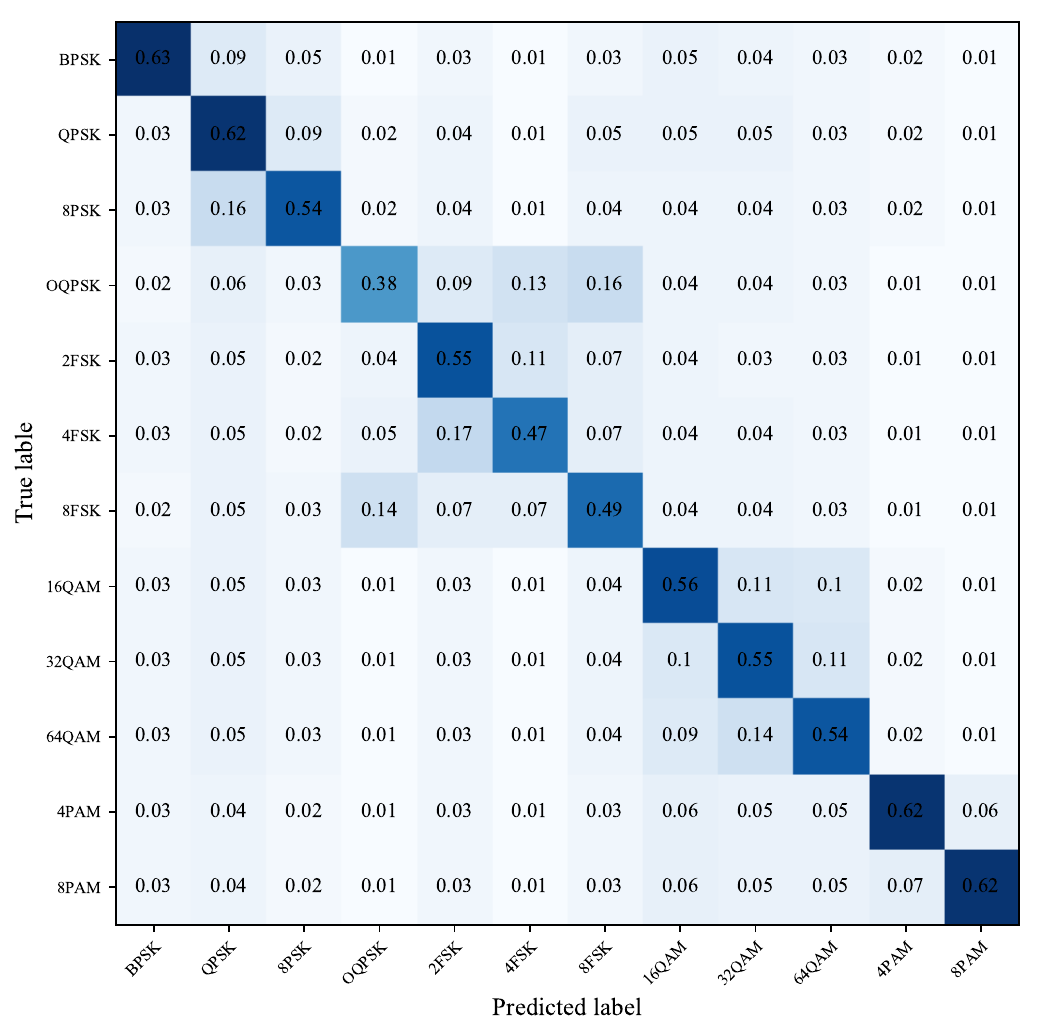}}
\subfigure[]{\label{fig:subfig1:sym5_hunxiao}
\includegraphics[width=0.32\linewidth]{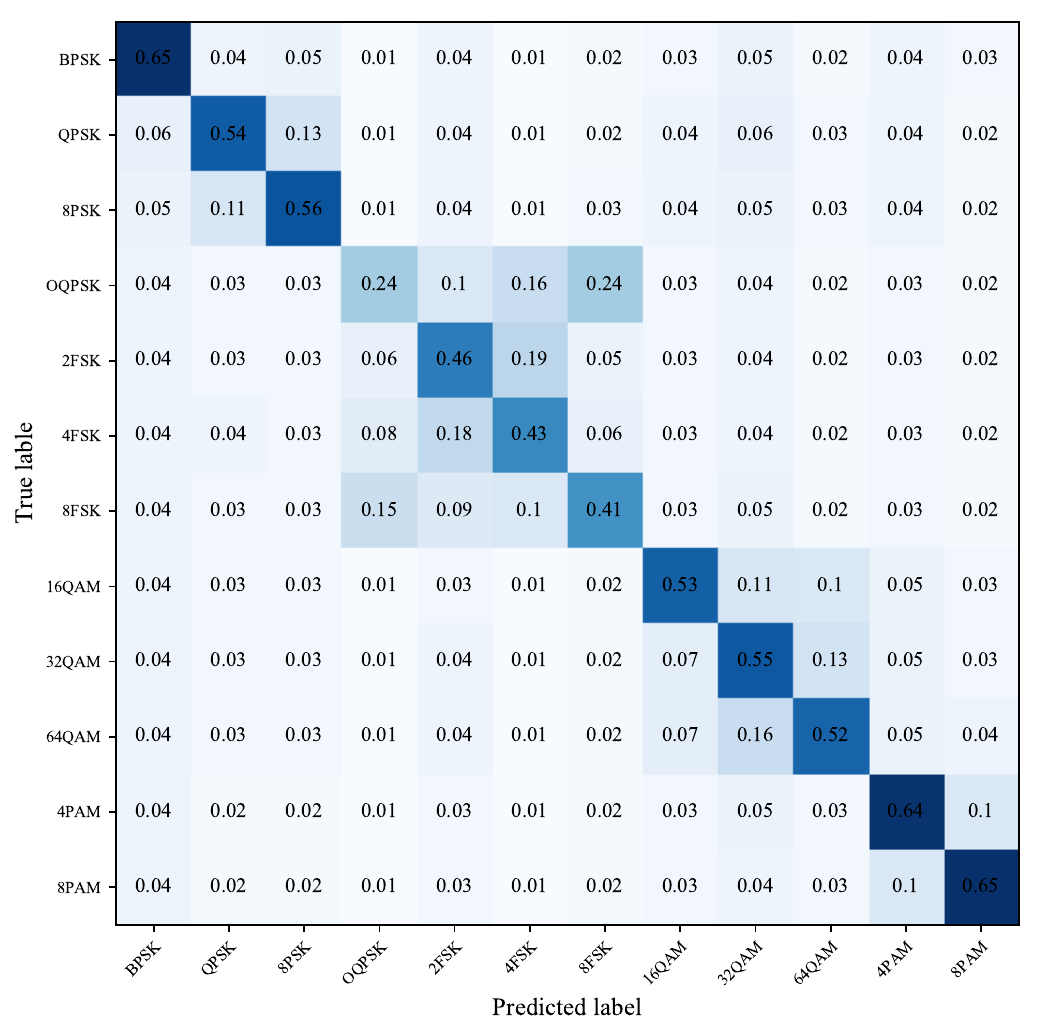}}
\label{fig:subfig1}
\centering
\end{figure*}

\subsection{Performance of Different Network Models}

\begin{figure}[t]
    \centering
    \includegraphics[width=7.5cm]{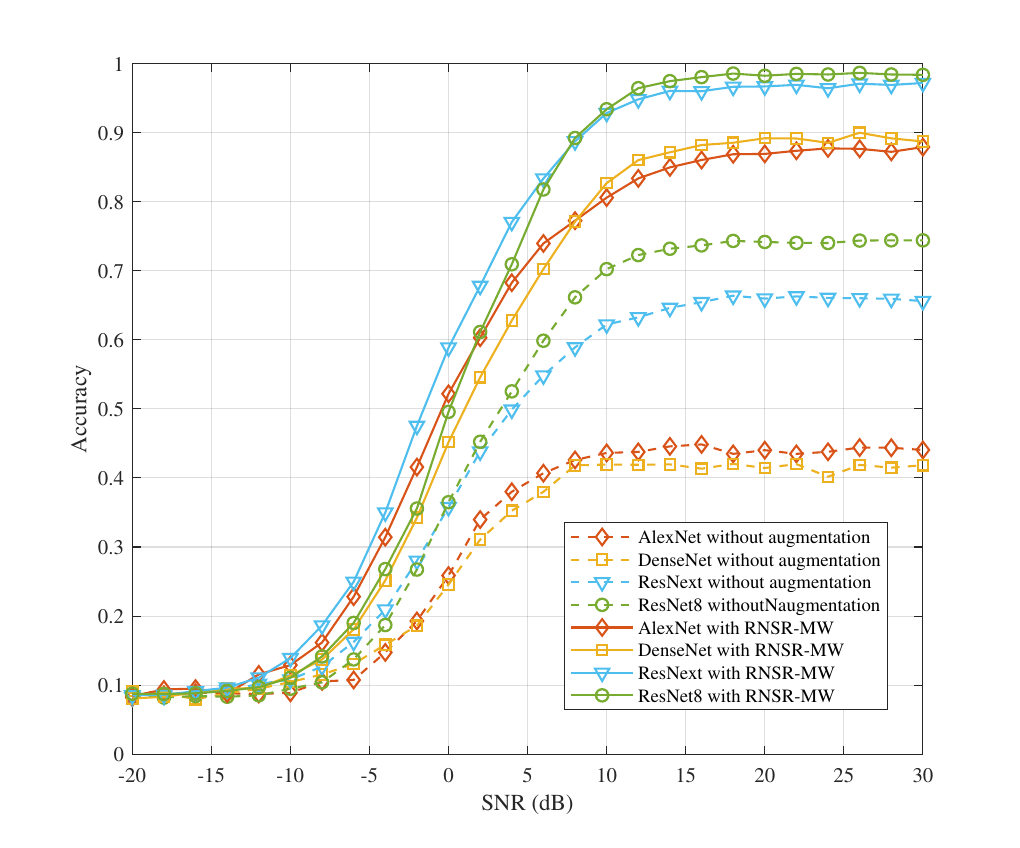}
    \caption{The performance of different network models at different SNRs.}
    \label{different_models}
\end{figure}

We use our proposed RNSR-MW method to conduct experiments on different network models, AlexNet \cite{2012ImageNet}, DenseNet \cite{8099726}, ResNext \cite{8100117}, ResNet8, and plot their recognition accuracy under different SNR in Fig .\ref{different_models}. The dataset we used is RML1024, and the number of times to perform wavelet transform is set to 3. In each of the four networks, the RNSR-MW augmentation method yields higher recognition accuracy than that of the method without augmentation, further affirming the effectiveness of our proposed augmentation method. Among the four networks, DenseNet has the lowest recognition accuracy. Notably, our proposed RNSR-MW method has the most significant performance gain on DenseNet and AlexNet, demonstrating its capability to augment low-performing networks. Moreover, the high performance achieved by both the without augmentation and RNSR-MW augmentation methods on ResNext and ResNet8 networks confirms the superiority of these two networks, which indicates that our proposed RNSR-MW augmentation method has better performance on convolutional networks with residuals.

\subsection{Comparison with Other Data Augmentation Methods}
In this scenario, we compare our proposed RNSR-MW augmentation method with several commonly used data augmentation methods, including the Flip-based method, SegCS-based method, and SegMC-based method. It is important to note that the Flip-based method involves both vertical and horizontal flipping, resulting in a total of four times the original number of samples. The SegCS-based method randomly divides each signal sequence into three segments with different lengths (denoted as SegCS3), while the SegMC-based method randomly divides each signal sequence into two segments with different lengths (denoted as SegMC2). In our proposed RNSR-MW method, the number of performing DWT we set is 3 and the number of replacement sequence generation we set is 1. Furthermore, in order to ensure that the dataset obtained by the augmentation method is consistent with our proposed RNSR-MW method, we set the number of random fragment partitioning times to 3 for the SegCS-based method (denoted as SegCS3$^3$). The network we used is ResNet8 and the dataset we used is RML1024. RNSR-MW method achieves a recognition accuracy of 60.574\%, which is significantly higher than the Flip-based, SegCS3-based, and SegMC2-based methods, which achieve accuracies of 47.318\%, 46.8\%, and 46.49\%, respectively. Compared to SegCS3$^3$ method with the same amount of samples, RNSR-MW method exhibits a remarkable improvement in accuracy, with nearly 8\% points of performance gain. All these results indicate the superiority of our proposed RNSR-MW augmentation method.

\begin{figure}[t]
    \centering
    \includegraphics[width=7.2cm]{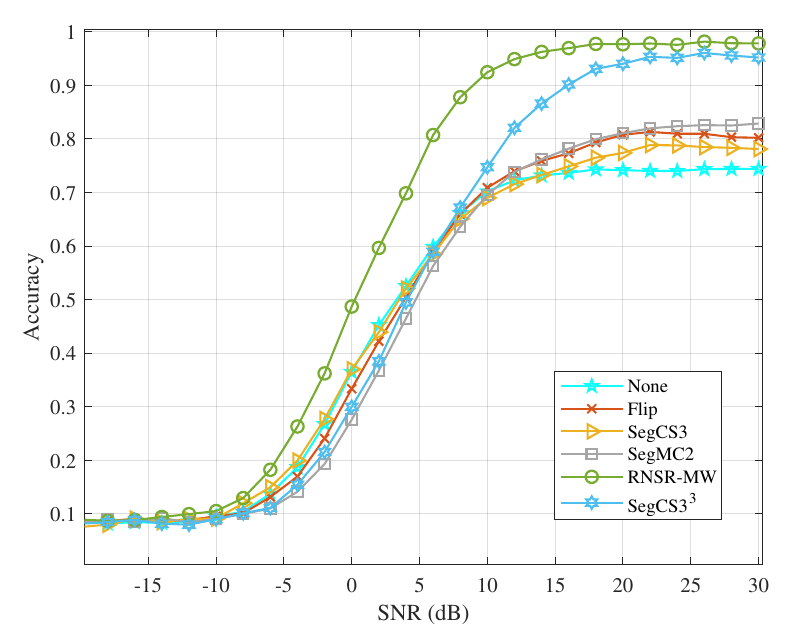}
    \caption{The performance of different data augmentation methods at different SNRs on RML1024 dataset.}
    \label{compare_method}
\end{figure}

We display the accuracy of these methods at various SNR in Fig. \ref{compare_method}, where none indicates that the raw IQ sequences have not been augmented for recognition. It can be inferred that both the Flip-based method and SegMC2-based method exhibit higher recognition accuracy compared to the SegCS3-based method, and there is a performance improvement in the high SNR of 12 dB to 30 dB compared to the method without augmentation. As the sample size increases, the performance of the SegCS3$^3$ method also greatly improves in the SNR range of 10 dB to 30 dB. It is very obvious that the RNSR-MW method we proposed is far superior to other augmented methods, and has significantly higher recognition accuracy than other methods in a wide range of SNR from $-$5 dB to 30 dB. 

\section{Conclusion}
We propose augmentation methods that replacing wavelet detail coefficients for reconstruction to generate new samples and expand the training set. Different generation methods are used to generate replacement sequences.
The simulation results indicate that our proposed augmentation methods significantly outperform the other augmentation methods, i.e., Flip-based, SegCS-based, and SegMC-based methods. And the performance on different networks further validates the universality of our proposed augmentation methods.
In the future, we will apply our proposed augmentation method to few-shot scenarios in other fields, such as signal detection, parameter estimation, and anomaly detection.




\ifCLASSOPTIONcaptionsoff
  \newpage
\fi



\bibliographystyle{IEEEtran}
\bibliography{main}


\end{document}